\documentclass[lettersize,journal]{IEEEtran}
\usepackage{amsmath,amsfonts}
\usepackage{algorithmic}
\usepackage{algorithm}
\usepackage{array}
\usepackage[caption=false,font=footnotesize,labelfont=sf,textfont=sf]{subfig}
\usepackage{textcomp}
\usepackage{stfloats}
\usepackage{url}
\usepackage{verbatim}
\usepackage{graphicx}
\usepackage{cite}
\usepackage{booktabs}
\usepackage{cleveref}
\usepackage{etoolbox}
\usepackage{booktabs}
\usepackage{makecell, multirow}
\usepackage{algorithmic}
\usepackage{algorithm}
\usepackage{tikz}
\usepackage{bm}
\usepackage{xspace}
\usepackage{amssymb}
\usepackage{amsthm}
\makeatletter
\DeclareRobustCommand\onedot{\futurelet\@let@token\@onedot}
\def\@onedot{\ifx\@let@token.\else.\null\fi\xspace}

\def\eg{\emph{e.g}\onedot} 
\def\ie{\emph{i.e}\onedot} 
 
\def\etc{\emph{etc}\onedot} 
 
\def\etal{\emph{et al}\onedot}
\makeatother

\newtheorem{theorem}{Theorem}

\begin{document}

\title{Disentangled Representation Learning with Transmitted Information Bottleneck}

\author{
  Zhuohang Dang, ~\IEEEmembership{}   
  Minnan Luo$^*$,~\IEEEmembership{}
  Chengyou Jia, ~\IEEEmembership{}
  Guang Dai,~\IEEEmembership{}
  Jihong Wang, ~\IEEEmembership{} \\
        Xiaojun Chang,~\IEEEmembership{}
        and~Jingdong Wang,~\IEEEmembership{}
\thanks{"Copyright © 2024 IEEE. Personal use of this material is permitted. However, permission to use this material for any other purposes must be obtained from the IEEE by sending an email to pubs-permissions@ieee.org."}
\thanks{$^*$Corresponding author: Minnan Luo.}
 \thanks{This work is supported by the National Nature Science Foundation of China (No. 62272374, No. 62192781, No. 62250009, No. 62137002), Natural Science Foundation of Shaanxi Province (No. 2024JC-JCQN-62), Project of China Knowledge Center for Engineering Science and Technology, Project of Chinese academy of engineering “The Online and Offline Mixed Educational Service System for ‘The Belt and Road’ Training in MOOC China”, and the K. C. Wong Education Foundation.}
  \thanks{Zhuohang Dang, Minnan Luo, Chengyou Jia and Jihong Wang are with the School of Computer Science and Technology, the Ministry of Education Key Laboratory of Intelligent Networks and Network Security, and the Shaanxi Province Key Laboratory of Big Data Knowledge Engineering, Xi’an Jiaotong University, Xi’an, Shaanxi 710049, China (e-mail: \{dangzhuohang,cp3jia,wang1946456505\}@stu.xjtu.edu.cn, minnluo@xjtu.edu.cn).}
	\thanks{Guang Dai is with the SGIT AI Laboratory, Xi’an 710048, China, and also with the State Grid Shaanxi Electric Power Company Ltd., State Grid Corporation of China, Xi’an 710048, China (e-mail: guang.gdai@gmail.com).}
	\thanks{Xiaojun Chang is with the School of Information Science and Technology, University of Science and Technology of China, Hefei 230026, China, and also with the Department of Computer Vision, Mohamed bin Zayed University of Artificial Intelligence (MBZUAI), Abu Dhabi, United Arab Emirates (e-mail: cxj273@gmail.com).}
        \thanks{Jingdong Wang is with Baidu Inc, Beijing 100085, China (e-mail: wangjingdong@outlook.com).}
}

\markboth{Journal of \LaTeX\ Class Files,~Vol.~14, No.~8, August~2021}
{Shell \MakeLowercase{\textit{et al.}}: A Sample Article Using IEEEtran.cls for IEEE Journals}

\maketitle

\begin{abstract}
  Encoding only the task-related information from the raw data, \ie, disentangled representation learning, can greatly contribute to the robustness and generalizability of models. 
  Although significant advances have been made by regularizing the information in representations with information theory, two major challenges remain: 
  1) the representation compression inevitably leads to performance drop;
  2) the disentanglement constraints on representations are in complicated optimization.
  To these issues, we introduce Bayesian networks with transmitted information to formulate the interaction among input and representations during disentanglement.
  Building upon this framework, we propose \textbf{DisTIB} (\textbf{T}ransmitted \textbf{I}nformation \textbf{B}ottleneck for \textbf{Dis}entangled representation learning), a novel objective that navigates the balance between information compression and preservation.
  We employ variational inference to derive a tractable estimation for DisTIB. This estimation can be simply optimized via standard gradient descent with a reparameterization trick.
  Moreover, we theoretically prove that DisTIB can achieve optimal disentanglement, underscoring its superior efficacy.
  To solidify our claims, we conduct extensive experiments on various downstream tasks to demonstrate the appealing efficacy of DisTIB and validate our theoretical analyses. 
  \end{abstract}

  \begin{IEEEkeywords}
    disentangled representation learning, information bottleneck, mutual information
    \end{IEEEkeywords}

    \section{Introduction}
    \IEEEPARstart{R}{epresentation} learning, a fundamental and significant topic within computer vision and artificial intelligence domains, aims to learn low-dimensional embeddings of raw data for easier exploitation \cite{goodfellow2016deep}.
    It serves as the cornerstone of various downstream tasks, such as classification \cite{deng2021mvf,hao2022attention}, detection \cite{fu2023learning,deng2024balanced} and generation \cite{tan2023attention,chen2024disendreamer}, \etc 
    Within this realm, traditional supervised learning approaches guide the learning of representations by leveraging corresponding labels, ensuring the learned representations effectively capture label-related parts of the raw data.
    However, a growing body of literature \cite{beery2018recognition,rosenfeldrisks,chen2022does,asgari2022masktune} has highlighted that only supervised learning may introduce spurious correlations between representations and labels, \eg, background bias \cite{tian2018eliminating}, thus greatly undermining the robustness and generalizability of representations.
    
    With the help of information theory, recent works have made significant contributions to capturing only label-related information, \ie, disentangled representation learning.
    These methods assume the information in raw data can be decomposed into two complementary parts: label-related and sample-exclusive, respectively. 
    The former maintains the discriminative information for classification (\eg, the digit content in MNIST \cite{deng2012mnist}), while the latter encodes all remaining label-irrelevant information (\eg, the location, size and writing style in MNIST).
    These methods mainly have two routes: regularization-based \cite{alemi2016deep,gao2021information,kolchinskycaveats,niu2013squared,yi2018neural} and disentanglement-based methods \cite{pan2021disentangled,qian2020unsupervised,zhao2021learning,hadad2018two,mathieu2016disentangling}. 
    Regularization-based methods attempt to add heuristic regularizations to the objective function, \eg, information constraints \cite{gao2021information,niu2013squared} and prior knowledge \cite{liu2019feature,yi2018neural,xu2022high}, \etc
    These regularizers aim to eliminate the redundant information as much as possible without performance loss, \ie, compression. 
    However, searching the performance-compression tradeoff with trivial regularization inevitably decreases the performance \cite{pan2021disentangled}. 
    Disentanglement-based methods propose novel training objectives \cite{pan2021disentangled,qian2020unsupervised} or strategies \cite{mathieu2016disentangling,hadad2018two} to fully disentangle the label-related and sample-exclusive information into separate representations. 
    In achieving this, these methods introduce separate variables to capture label-related and sample-exclusive information, together with disentanglement constraints to eliminate information overlap between these variables.
    Despite their efficacy on realistic datasets, these methods usually employ complex strategies to optimize the disentanglement constraints, such as adversarial training \cite{mathieu2016disentangling}, two-step architecture \cite{hadad2018two} and objective factorization \cite{hwang2020variational}, \etc
    Consequently, they suffer from the unstable and inefficient optimization process \cite{creswell2018generative}.

    To overcome the aforementioned limitations, we suggest exploring a novel information-theoretic objective to enhance the effectiveness of disentangled representation learning.
    Specifically, we leverage two Bayesian networks to formulate the variable interactions in disentangled representation learning, aligning with the principles of information compression and preservation in information theory.
    Facilitated by these Bayesian networks, we formalize the transmission of information among variables, which can be optimized to ensure both the compactness and informativeness of the learned representations.
    Next, we delve into the balance between representation compactness and informativeness, culminating in the introduction of a novel objective based on the information bottleneck principle: \textbf{T}ransmitted \textbf{I}nformation \textbf{B}ottleneck For \textbf{Dis}entangled Representation Learning (DisTIB).
    We then derive a tractable variational approximation for DisTIB, which can be stably optimized.
    Moreover, we conduct an in-depth theoretical analysis of our DisTIB, in which we prove its convergence to optimal disentanglement and thus avoiding the notorious performance-compression trade-off encountered in prior methods.
    These theoretical results demonstrate the superiority of DisTIB in stability and efficacy. 
    We empirically demonstrate the efficacy of DisTIB on various downstream tasks, including adversarial robustness, supervised disentangling and more challenging few-shot learning, \etc
    
    Our contributions are as follows:
    \begin{itemize}
        \item By formulating the variable interactions with Bayesian networks and transmitted information, we propose a novel transmitted information bottleneck to disentangle the label-related and sample-exclusive information, whose objective can be stably optimized with standard variational estimation.
        \item We provide an in-depth theoretical analysis of our DisTIB, in which we prove that it can achieve representations with optimal disentanglement, thereby avoiding the notorious performance-compression trade-off.
        \item Extensive superior experiment results on adversarial robustness, supervised disentangling and few-shot learning, \etc, reveal the effectiveness of DisTIB, and the validity of our theoretical analyses.
    \end{itemize}
    
        The rest of the paper unfolds as follows: Section \uppercase\expandafter{\romannumeral2} introduces related works, notably the disentangled representation learning and information bottleneck. Section \uppercase\expandafter{\romannumeral3} delves into the details of the proposed DisTIB, while Section \uppercase\expandafter{\romannumeral4} focuses on its optimization. In Section \uppercase\expandafter{\romannumeral5}, we present the datasets, implementation details, and experimental results, including comparisons with state-of-the-art methods and ablation studies. Section \uppercase\expandafter{\romannumeral6} concludes the paper and outlines avenues for future research.
    
    \section{Related Work}
    \subsection{Disentangled Representation Learning}
    Disentangled representation learning \cite{huang2021learning,pan2023disentangled,li2022disentangled1} aims to recover the mutually independent factors from data. For example, $\beta$-VAE \cite{higgins2016beta} learns the factorized representations by controlling the trade-off between the reconstruction and prediction. Hadad \etal \cite{hadad2018two} propose a two-step framework that extracts label-related and sample-exclusive information sequentially.
    Information theory is also introduced to regularize information in disentangled representations. InfoGAN \cite{chen2016infogan} facilitates disentangled representation learning by maximizing the mutual information between input and latent. SPEECHSPLIT \cite{qian2020unsupervised} leverages three information bottlenecks on different encoders to disentangle speech components. IIAE \cite{hwang2020variational} proposes a novel training objective that combines the conventional evidence lower bound and disentanglement regularizations to extract the domain shared and exclusive information. FactorVAE \cite{kim2018disentangling} and TCVAE \cite{chen2018isolating} factorize the training objective of the traditional variational encoder to directly penalize the total correlation term for better disentanglement.
    Some work \cite{zhao2021learning,niu2013squared,dang2023counterfactual} introduce mutual information regularization to better maintain the label-related information.
    Nevertheless, these methods involve complicated optimization strategies, leading to unstable and inefficient optimization procedures.
    In contrast, our DisTIB can be stably optimized with variational inference, and is proved to exactly control the information amount in representations to ensure optimal disentanglement.

    \subsection{Information Bottleneck}
    Information bottleneck \cite{tishby2000information,yang2023individual} is introduced to find effective, highly compact features, compressing the information from raw data while only maintaining the information on the label. 
    This is achieved by searching the trade-off between two mutual information (MI) terms: 1) maximizing the MI between features and target, \ie, prediction; 2) minimizing the MI between features and raw data, \ie, compression.
    Recently, VIB \cite{alemi2016deep} has employed variational inference to derive tractable estimations for the MI terms, making its optimization easier when applied to deep neural networks. 
    Tishby \cite{tishby2015deep} leverages the IB principle to quantify the MI between layers in DNN and gives an information theoretic limit of DNN under finite samples.
    Some works extend the idea of IB into various downstream tasks, including multi-view learning \cite{federici2020learning}, graph representation learning \cite{wu2020graph} and domain generalization \cite{du2020learning}, \etc
    In this paper, we introduce the novel Transmitted Information Bottleneck for Disentangled Representation Learning. Additionally, we provide in-depth theoretical proof of its disentanglement efficacy, confirming the robustness and generalizability of learned representations.

    \section{Methodology}

In this section, we begin by defining disentangled representation learning, followed by modeling variable interactions using Bayesian networks.
We then formulate DisTIB and provide an in-depth theoretical insight.
\subsection{Transmitted Information Bottleneck}
\subsubsection{Problem Definition}
Consider a set of paired data sampled from the joint distribution $(x,y) \thicksim p(x,y)$, where $x\in X$ is extracted from the observational data and $y\in Y$ is its corresponding label, respectively. 
Similar to \cite{yue2021counterfactual,pan2021disentangled}, we assume that the observational data $X$ exhibits sample-specific factors of variations while sharing some common factors of variations within a class. 
For example, samples in the MNIST dataset can share the same semantic content (\ie, digit), while having different styles (\ie, writing, position, and size, \etc).
Given this data, our goal of disentangled representation learning is to seek a pair of complementary structured representations: label-related information $A$ that captures the discriminative characteristics of label $Y$, and the sample-exclusive information $Z$ that encodes all label-irrelevant factors. 
Notably, our disentanglement is group-level, eschewing a focus on the specific semantics of each dimension since it requires enormous external knowledge.

\subsubsection{Variable Bayesian Networks}
\begin{figure}[!t]
    \centering
    \subfloat[Information Compression.]{{\includegraphics[width=0.42\linewidth]{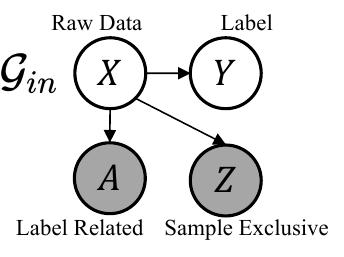}\label{fig: G_in}}}
    \hspace{5mm}
    \subfloat[Information Preservation.]{{\includegraphics[width=0.42\linewidth]{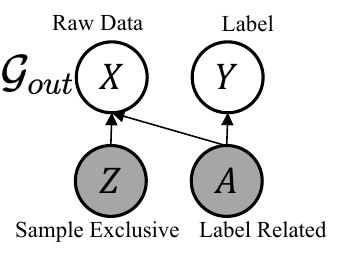}\label{fig: G_out}}}
    \caption{
    Illustration of Bayesian networks for information compression and preservation. The white and gray nodes denote input and target disentangled variables, respectively. The arrow indicates the direction of information flow, \ie, variable interactions, during the disentangled representation learning.
    }
    \label{fig: G}
\end{figure}

In the framework of disentangled representation learning, it's pivotal to understand the mechanisms behind extracting and preserving representations from raw data $x$.
To this end, we focus on two core principles of information theory: 1) information compression that guarantees the representation compactness, 2) information preservation for ensuring the representation informativeness to target variables.
Similar to \cite{slonim2006multivariate,hu2020dmib,zhang2020learning}, in \Cref{fig: G}, we leverage two Bayesian networks $\mathcal{G}_{in}$ and $\mathcal{G}_{out}$ to formulate the graphical model encoding these two structures, where vertices are annotated by names of random variables. 
Specifically, in the information compression stage, we aim to encode disentangled label-related information $A$ and sample-exclusive information $Z$ from the original input $X$.
Subsequently, $\mathcal{G}_{in}$ identifies variables $X$ as root while $\{A,Z\}$ as leaves, where edges denote the compression relation among variables.
	On the other hand, in the information preservation stage, we aim to ensure that the label-related information $A$ captures the discriminative characteristics of the label $Y$ (edge $A \rightarrow Y$); the sample-exclusive information $Z$ encodes all label-irrelevant factors. Furthermore, $A$ and $Z$ should be complementary, thereby collectively encoding all input information (edges $A \rightarrow X$ and $Z \rightarrow X$).
    As a result, $\mathcal{G}_{out}$ defines variables they should preserve information about as leaves, indicating the information preservation relation among variables. 
These two Bayesian networks offer a holistic perspective of the intricate balance between compressing and preserving information.

\subsubsection{Formulation}
The Bayesian network $\mathcal{G}_{in}$ encodes the conditional independence among disentanglement variables, \eg, $A \perp\!\!\!\perp Z \mid X$.
Let $p^{\mathcal{G}_{in}}(X,Y,A,Z)$ denote the joint distribution of variables encoding the dependencies implied in $\mathcal{G}_{in}$, the information variables shared about each other in compression stage can be formulated as:
\begin{equation}
    I^{\mathcal{G}_{in}}=D_{K L}[p^{\mathcal{G}_{in}}(X,Y,A,Z) \| p(X)p(Y)p(A)p(Z)],
\end{equation}
where $D_{KL}$ refers to KL divergence and $I^{\mathcal{G}_{in}}$ is the multivariate extension of conventional mutual information, namely multi-information \cite{studeny1998multiinformation}.
As in \cite{slonim2006multivariate}, minimizing $I^{\mathcal{G}_{in}}$ ensures that $A$ and $Z$ capture the information from $X$ while maintaining their mutual disentanglement, corresponding to the desired information compression process.
However, the information compression process primarily emphasizes representation compactness, while overlooking the informativeness of the representation, \ie, \emph{Do disentanglement variables capture the target information from original data?}
To address this, we further employ $\mathcal{G}_{out}$ that encodes the informativeness dependencies between disentanglement variables and observational data, \eg, $X \perp\!\!\!\perp Y \mid A$.
Similarly, let $p^{\mathcal{G}_{out}}(X,Y,A,Z)$ denote the joint distribution of variables encoding the dependencies implied in $\mathcal{G}_{out}$. 
By formulating $I^{\mathcal{G}_{out}}$ that denotes the information variables contained about each other in the preservation stage, we follow \cite{mcgill1954multivariate} to maximize it for encouraging the representation informativeness.
Finally, inspired by the IB principle, we aim to achieve the optimal disentanglement of representations by the following objective:
\begin{equation}
    \min_{A,Z} L_{DisTIB}= -I^{\mathcal{G}_{out}} + \beta I^{\mathcal{G}_{in}}.
    \label{eq: IB}
\end{equation}
The hyperparameter $\beta\in[0,1]$ explores the trade-off between representation compactness and informativeness, akin to conventional IB \cite{alemi2016deep}.
We then quantify the multi-information in \Cref{eq: IB} with the following \Cref{MI quant}.
\begin{theorem}
Given Bayesian networks $\mathcal{G}$ over variables $\bm{\mathcal{X}}=\{X_1,\cdots,X_n\}$, let $\mathbf{P a}_{X_i}^\mathcal{G}$ denotes parents of $X_i$ in $\mathcal{G}$, the joint distribution can be factorized as $p(\bm{\mathcal{X}})=\Pi_{i=1}^n p(X_i | \mathbf{P a}_{X_i}^\mathcal{G})$, thus the multi-information $\mathcal{I}^\mathcal{G}$ can be quantified by
\begin{align}
    \mathcal{I}^\mathcal{G} & =D_{K L}[p\left(X_1, \ldots, X_n\right) \| p\left(X_1\right) \ldots p\left(X_n\right)] \notag\\
    & = D_{K L}\left[\Pi_{i=1}^n p(X_i \mid \mathbf{P a}_{X_i}^\mathcal{G}) \| p\left(X_1\right) \ldots p\left(X_n\right)\right] \notag\\
    & =\sum_i I(X_i;\mathbf{P a}_{X_i}^\mathcal{G}),
\end{align}
where $I(X_i;\mathbf{P a}_{X_i}^\mathcal{G})$ is the transmitted information, an extension of mutual information that measures the information transmitted from multi-source $\mathbf{P a}_{X_i}^\mathcal{G}$ to receiver $X_i$.
\begin{proof}
    Please see supplementary for detailed proof.
\end{proof}
\label{MI quant}
\end{theorem}
Subsequently, according to \Cref{MI quant} and the Bayesian networks $\mathcal{G}_{in}$ and $\mathcal{G}_{out}$ defined for information compression and preservation, DisTIB is reduced to an IB-based trade-off among the transmitted information \cite{mcgill1954multivariate}, resulting in the formulation of \emph{transmitted information bottleneck}, \ie,
\begin{align}
\label{eq: graph}
    &\min_{A,Z} L_{DisTIB} =  -I^{\mathcal{G}_{out}} + \beta I^{\mathcal{G}_{in}} \\
    &=-\sum_{X_i\in \mathcal{G}_{out}} I(X_i;\mathbf{P a}_{X_i}^{\mathcal{G}_{out}}) + \beta \sum_{X_i\in \mathcal{G}_{in}} I(X_i;\mathbf{P a}_{X_i}^{\mathcal{G}_{in}})\notag.
\end{align}
More specifically, by incorporating the graphical structures of $\mathcal{G}_{in}$ and $\mathcal{G}_{out}$ from \Cref{fig: G} into \Cref{eq: graph}, we derive the empirical objective of DisTIB as:
\begin{align}
\label{eq: origin}
    \min_{A,Z} L_{DisTIB} 
    =-&[\underbrace{I(X;A,Z)}_{Sufficiency}+\underbrace{I(A;Y)}_{Prediction}]\\
    +&\beta\underbrace{[I(X;A)+I(X;Z)+I(X;Y)]}_{Compression},\notag
\end{align}
where the compression term indicates the learning process of the disentangled variables, \ie, the model encodes the information of $X$ into $A$, $Z$ and $Y$, respectively.
Moreover, maximizing the sufficiency term $I(X;A,Z)$ encourages the disentangled representation pair $(A,Z)$ to contain all information about $X$, \ie, $(A,Z)$ can restore $X$ losslessly.
For the prediction term, maximizing $I(A;Y)$ encourages $A$ to preserve information in $X$ about $Y$, \ie, label-related information, as much as possible.

In this paper, we analyze the scenario where the relationship between $Y$ and observational data $X$ is deterministic \cite{rodriguez2020convex}, \ie, there exists a surjective, injective, or bijective function \(f\) such that \(f(X)=Y\).
This assumption is prevalent, covering a wide range of downstream applications. For example, the input maintains definitive class labels and locations in classification and detection tasks, or the output image contains desired input label conditions in controllable image generation, showcasing a deterministic input-label relationship.
In this sense, $I(X;Y) = H(Y)$ is an ignorable constant in optimization, where $H(Y)$ is the entropy of $Y$ specified by the dataset.
This constraint, $I(X;Y) = H(Y)$, offers foundations for analyzing DisTIB's disentanglement efficacy in \Cref{the2}.

\subsection{Theoretical Analyses}
In this section, we conduct an in-depth theoretical analysis to validate the disentanglement effectiveness of DisTIB in the following three aspects: completeness, mutual disentanglement and minimal sufficiency of $A$. 
\begin{theorem}
\label{the2}
    Given data $X$ with label $Y$, let $A$ and $Z$ represent the label-related and sample-exclusive information, respectively. Let $A^*$ and $Z^*$ signify the optimal label-related and sample-exclusive information obtained by minimizing $L_{DisTIB}$, where $L_{DisTIB}^*$ is the global minimum of $L_{DisTIB}$. 
    Then for any given $\epsilon>0$, there exists a $\delta>0$ such that $L_{DisTIB}-L^*_{DisTIB}<\delta$ and we have:
    \begin{itemize}
        \item \textbf{completeness:} The optimal label-related representation $A^*$ and sample-exclusive representation $Z^*$ are sufficient for data $X$, i.e., $|H(X)-I(X;A^*,Z^*)|<\epsilon$.

        \item \textbf{mutual disentanglement:} The optimal label-related representation $A^*$ and sample-exclusive representation $Z^*$ are disentangled, i.e., $|I(A^*;Z^*)|<\epsilon$.

        \item \textbf{minimal sufficiency of $A$:} Given $I(X;A)\leq I(X;Y)$, the optimal label-related representation $A^*$ is minimal sufficient, i.e., $|I(Y;A^*)-H(Y)|+|I(X;A^*|Y)|<\epsilon$.
    \end{itemize}
\label{th: 1}
    \begin{proof}
        Please see supplementary for detailed proof.
    \end{proof}
\end{theorem}
Theorem \ref{the2} reveals that the proposed DisTIB can achieve the optimal disentanglement between $A$ and $Z$.
Specifically, the completeness indicates that the disentangled representation pair $(A,Z)$ captures all input information of $X$, while the mutual disentanglement denotes that there is no overlap between $A$ and $Z$.
On the other hand, the minimal sufficiency of $A$ indicates that $A$ captures all label-related information without extra sample-exclusive noise.
Moreover, since $I(X;Z)=H(X)-H(X|Z)$ has no analytic solution, \Cref{th: 1} does not impose any specific constraint on the information amount contained within $Z$.
However, given \Cref{th: 1}, the aforementioned three constraints collaboratively lead $Z$ to capture sample-specific information without label-related noise, \ie, optimal disentanglement of $Z$.
In the following, we delve deeper into the disentanglement efficacy of DisTIB for a more thorough understanding.

\begin{figure}[!t]
    \centering
    \includegraphics[width=1\linewidth]{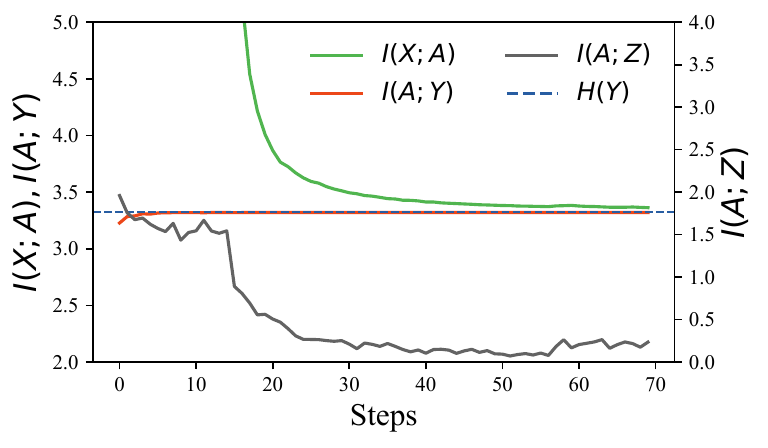}
\caption{
Illustration of DisTIB's training procedure on MNIST dataset, where disentangled variables gradually converge to their optimal value. In detail, the solid lines represent the estimation of mutual information between representations. The blue dashed line represents entropy of labels, $H(Y)$, which is the optimal value for the disentangled representation mutual information $I(X;A)$ and $I(A;Y)$.
}
    \label{fig: train_log}
\end{figure}

Compared to previous regularization-based methods \cite{kolchinsky2019nonlinear,alemi2016deep,niu2013squared}, DisTIB can avoid the notorious performance-compression trade-off. Specifically, \cite{pan2021disentangled} proves that previous regularization techniques inevitably decrease performance as the compression level intensifies. Differently, \Cref{fig: train_log} demonstrates that as the DisTIB training process progresses, the disentangled variables gradually converge to their optimal values, further validating the theoretical analysis of DisTIB's disentanglement performance in \Cref{th: 1}.
On the other hand, although previous disentanglement-based methods employ explicit disentanglement constraints, \ie, $I(A,Z)$, to directly disentangle features with apparent simplicity, they suffer from training instability and inefficiency due to their complex optimization strategies, \eg, adversarial training \cite{pan2021disentangled}, objective factorization \cite{hwang2020variational}, separate optimization \cite{sanchez2020learning}, \etc. Conversely, DisTIB can be estimated with variational inference, resulting in a more stable and concise optimization process.

\section{DisTIB Optimization and Inference}
\label{Optimization}
In the previous section, we propose DisTIB to optimally disentangle label-related and sample-exclusive information.
However, it is intractable to directly optimize \Cref{eq: origin}, since the transmitted information typically consists of integrals on high-dimensional space. 
Therefore, similar to the previous work \cite{alemi2016deep}, we employ variational inference to derive tractable bounds for these terms. The derivation details are shown in supplementary due to the page limits.

\subsection{Compression Term}
To minimize $I(X;Z)$, we approximate the marginal distribution $p(Z)$ with the parameterized distribution $r(Z)$.
In this sense, the variational upper bound for $I(X;Z)$ is formulated as:
\begin{equation}
\begin{aligned}
    I(X;Z)  &\leq \int p(X,Z)\log\frac{p(Z|X)}{r(Z)}\,dx \,dz\\
  &=\mathbb{E}_{P(X)}[D_{KL}(p(Z|X)\| r(Z))].
\label{eq: kl_div}
\end{aligned}
\end{equation}
Similarly, $I(X;A)$ can be estimated by using variational distribution $r(A)$ (see supplementary materials for details).

\subsection{Sufficiency Term}
For the maximization of $I(X;A,Z)$, we leverage the parameterized posterior distribution $q_\phi(X|A,Z)$ to approximate the true posterior distribution $p(X|A,Z)$, where $\phi$ are the distribution parameters. The tractable lower bound of $I(X;A,Z)$ can be derived as
{\small
\begin{equation}
\begin{aligned}
    &I(X;A,Z) =\int p(X,A,Z)\log\frac{p(X|A,Z)}{p(X)} \,da\,dx\,dz\\
      &\geq \int p(X,A,Z)\log q_{\phi}(X|A,Z) \,da\,dx\,dz+H(X),
\end{aligned}
\end{equation}}where $H(X)$ is an ignorable constant. 
Moreover, this estimation allows us to implement the sufficiency term as a generator based on disentangled representations $(A,Z)$, enabling DisTIB with end-to-end disentangled generation ability.

\definecolor{commentcolor}{RGB}{110,154,155}  
\newcommand{\PyComment}[1]{\ttfamily\textcolor{commentcolor}{\# #1}}
\newcommand{\PyCode}[1]{\ttfamily\textcolor{black}{#1}}
\begin{algorithm}[t]\scriptsize
            \PyComment{$Enc_Z, Enc_A$: label-irrelevant and -related encoder} \\
            \PyComment{$G_{\Phi}$: generators based on disentangled features (A,Z)}\\
                \PyCode{def inference(input,label):}
                    \PyComment{input:(N,C); label:(N,1)}\\
                    \hspace*{4mm} \PyComment{extract disentangled features}\\
                    \hspace*{4mm} \PyCode{A = $Enc_A$.forward(input)}\PyComment{N*dim\_A}\\
                    \hspace*{4mm} \PyCode{Z = $Enc_Z$.forward(input)}\PyComment{N*dim\_Z}\\
                    \hspace*{4mm} \newline
                    \hspace*{4mm} \PyComment{generations on different (A,Z) combination}\\
                    \hspace*{4mm} \PyComment{dimension expansion: (N*N)*dim\_A, (N*N)*dim\_Z}\\
                    \hspace*{4mm} \PyCode{A\_1 = A.repeat(N,1)}\\
                    \hspace*{4mm} \PyCode{Z\_1 = Z.repeat\_interleave(N,0)}\\
                    \hspace*{4mm} \PyCode{gen = $G_{\Phi}$.forward(A\_1, Z\_1)}\PyComment{(N*N)*C}\\
                    \hspace*{4mm} \PyComment{determined by A, with shape (N*N)*1}\\
                    \hspace*{4mm} \PyCode{gen\_label = label.repeat(N,1)}\\
                    \hspace*{4mm} \newline
                    \hspace*{4mm} \PyComment{label-related information, generations, labels}\\
                    \hspace*{4mm} \PyCode{return A,(gen,gen\_label)}
        \caption{Pseudo-code for disentangled representation extraction and sample generation with DisTIB.}
        \label{alg: inference}
        \end{algorithm}

        \begin{table*}[ht]
		\centering
  \caption{Generalization and adversarial robustness performance (\%) on MNIST, FashionMNIST and CIFAR10 datasets. For fair comparisons, we employ the same model architecture and directly report the corresponding results of \cite{pan2021disentangled}, where the hyperparameters have been carefully tuned for best efficacy.}
		\label{table_generalization_attack}
		\begin{tabular}{cccccccccc}
			\toprule
			\multicolumn{2}{c}{MNIST} & VIB \cite{alemi2016deep} & NIB \cite{kolchinsky2019nonlinear} & s-NIB \cite{rodriguez2020convex} & SNIB \cite{yang2023snib} & ExpIB \cite{wu2023exponential}  & DisenIB \cite{pan2021disentangled} & DisGenIB \cite{dang2024disentangled} & \textbf{Ours} \\
			\midrule
			\multicolumn{2}{c}{Generalization} & 97.6 & 97.2 & 93.3 & 93.5 & 97.4 & 98.2 & 98.7 & \textbf{98.8} \\
			\midrule
			\multirow{3}{*}{\shortstack{Adversary \\ Robustness}} & $\epsilon=0.1$ & 74.1 / 73.4 &  75.2 / 75.2 & 61.3 / 62.0 & 54.3 / 54.7 & 74.7 / 72.8 & 94.3 / 90.2 & 97.4 / 92.5 & \textbf{98.7 / 94.8} \\
			& $\epsilon=0.2$ & 19.1 / 20.8  & 21.8 / 23.6 & 24.1 / 24.5 & 13.6 / 14.4 & 23.1 / 24.3 & 81.5 / 80.0 & 85.3 / 82.6 & \textbf{87.9 / 85.5} \\
			& $\epsilon=0.3$ & 3.5 / 4.2  & 3.2 / 3.4 & 9.3 / 9.9 & 3.8 / 3.8 & 5.4 / 5.8  & 68.4 / 67.8 & 70.5 / 69.4 & \textbf{74.8 / 74.4} \\
			\bottomrule
		\end{tabular}
		\\
		\begin{tabular}{cccccccccc}
			\toprule
			\multicolumn{2}{c}{FashionMNIST} & VIB \cite{alemi2016deep} & NIB \cite{kolchinsky2019nonlinear} & s-NIB \cite{rodriguez2020convex} & SNIB \cite{yang2023snib} & ExpIB \cite{wu2023exponential}  & DisenIB \cite{pan2021disentangled} & DisGenIB \cite{dang2024disentangled} & \textbf{Ours} \\
			\midrule
			\multicolumn{2}{c}{Generalization} & 90.1 & 89.7 & 89.6 & 84.6 & 88.7 & 90.1 & 89.8 & \textbf{91.7} \\
			\midrule
			\multirow{3}{*}{\shortstack{Adversary \\ Robustness}} & $\epsilon=0.1$ & 26.3 / 29.0 &  26.7 / 28.7 & 28.0 / 29.4 & 22.4 / 23.5 & 25.4 / 27.6 & 59.6 / 62.0 & 53.4 / 58.7 & \textbf{62.5 / 65.1} \\
			& $\epsilon=0.2$ & 3.7 / 4.4  & 2.6 / 0.3 & 0.7 / 0.8 & 3.5 / 4.2 & 4.5 / 5.2 & 45.5 / 48.4 & 41.3 / 44.2 & \textbf{47.4 / 50.1} \\
			& $\epsilon=0.3$ & 0.1 / 0.1  & 0.1 / 0.1 & 0.3 / 0.4 & 0.7 / 0.6 & 0.7 / 0.8 & 32.5 / 34.9 & 28.7 / 31.7 & \textbf{35.6 / 37.4} \\
			\bottomrule
		\end{tabular}
		\\
		\begin{tabular}{cccccccccc}
			\toprule
			\multicolumn{2}{c}{CIFAR10} & VIB \cite{alemi2016deep} & NIB \cite{kolchinsky2019nonlinear} & s-NIB \cite{rodriguez2020convex} & SNIB \cite{yang2023snib} & ExpIB \cite{wu2023exponential}  & DisenIB \cite{pan2021disentangled} & DisGenIB \cite{dang2024disentangled} & \textbf{Ours} \\
			\midrule
			\multicolumn{2}{c}{Generalization} & 97.9 & 97.3 & 93.7 & 94.5 & 93.1 & \textbf{98.8} & 97.5 & 98.1 \\
			\midrule
			\multirow{3}{*}{\shortstack{Adversary \\ Robustness}} & $\epsilon=0.1$ & 64.3 / 62.4 &  69.2 / 65.2 & 62.3 / 61.0 & 51.0 / 49.2 & 61.7 / 60.2  & 70.4 / 70.3 & 68.7 / 68.2 & \textbf{73.9 / 73.2} \\
			& $\epsilon=0.2$ & 18.1 / 19.8  & 22.8 / 23.6 & 27.1 / 25.2 & 18.4 / 18.6 & 22.3 / 20.9  & 49.2 / 48.7 & 48.3 / 47.6 & \textbf{52.4 / 51.3} \\
			& $\epsilon=0.3$ & 2.0 / 3.2  & 5.2 / 4.3 & 10.1 / 10.7 & 4.7 / 4.3 & 8.4 / 8.1 & 29.3 / 30.0 & 28.5 / 28.4 & \textbf{32.4 / 31.9} \\
			\bottomrule
		\end{tabular}
	\end{table*}

\subsection{Prediction Term}
Let the variational distribution $q_\gamma(Y|A)$ with parameters $\gamma$ be a variational approximation of true posterior distribution $p(Y|A)$. The corresponding tractable lower bound of $I(A;Y)$ is formulated as
\begin{equation}\label{eq:I(A;Y)}
    \begin{aligned}
        I(A;Y) &= \int p(A,Y)\log \frac{p(Y|A)}{p(Y)} \,da\,dy \\
        &\geq\int p(A,Y)\log q_\gamma(Y|A) \,da\,dy + H(Y).
    \end{aligned}
\end{equation}

The estimations above give a tractable bound for minimizing DisTIB on the basis of standard mini-batch gradient descent with the reparameterization trick, demonstrating the stability of DisTIB's optimization.

\subsection{Model Inference}
\Cref{alg: inference} shows the pseudo-code of DisTIB's inference process, where $N$ is the batchsize; $C_1$ and $C_2$ are feature dimensions of label-related and sample-exclusive representation. The usage of inference output is task-dependent, \eg, $A$ for disentangled representation learning, while $(gen,gen\_label)$ for disentangled sample generation.

\section{Experiments}
In this section, we conduct extensive experimental studies to evaluate our DisTIB. We first introduce our experimental settings and then discuss the results of the experiments. Specifically, we aim to answer the following questions:
\begin{itemize}
    \item \textbf{RQ1:} Can our DisTIB achieve superior performance on tasks inherently reliant on the informativeness and compactness of representations \cite{goodfellow2014explaining,snell2017prototypical}, such as adversarial robustness, generalization, supervised disentangling and few-shot learning? 
    \item \textbf{RQ2:} Whether the proposed DisTIB achieves the optimal disentanglement?
    \item \textbf{RQ3:} How does each component of DisTIB facilitate the disentanglement?
\end{itemize}

\subsection{Implementation Details}
Similar to \cite{alemi2016deep,pan2021disentangled,kolchinsky2019nonlinear}, in DisTIB, all encoders and decoders are implemented using the Deep Gaussian family with LeakyReLU as the activation function, where the dimension of the hidden layer to 4096. In the following, we give details of the corresponding modules.
      \begin{enumerate}
        \item We employ two stochastic encoders, $p(A|X)$ and $p(Z|X)$, to respectively capture label-relevant and sample-specific information from the input. 
          Each encoder is formulated by the form $N(f_{\mu_{*}}(X),f_{\sigma_{*}^2}(X))$ and $*\in \{A,Z\}$.
          Specifically, we employ distinct neural networks to predict the mean $f_{\mu_{*}}(X)$ and variance $f_{\sigma_{*}^2}(X)$ of corresponding features, where $f_{\mu_{*}}(X),f_{\sigma_{*}^2}(X)$ keeps identical architecture as three-layer fully-connected networks.
          Additionally, their input size aligns with the feature dimensions, while their output is specifically tailored to the respective sizes of label-related and sample-specific information.
        \item The generator $q_{\phi}(X|A,Z)$ is designed to generate samples based on disentangled feature pairs $(A,Z)$. It has the form $N(f_{\mu_{\phi}}(A,Z),I)$ whose variance is fixed and mean is generated from a three-layer fully-connected network $f_{\mu_{\phi}}(A,Z)$. The input dimensions are determined by the combined sizes of label-related and sample-specific information, whereas the output retains the original feature size.
        \item The decoder $q_{\gamma}(Y|A)$ is of the form $q(A)=softmax(AW+b)$, which is a simple classification model to predicts the label $Y$ based on label-related information $A$.
      \end{enumerate}
Furthermore, the mutual information objective $\mathcal{L}_{DisTIB}$ is implemented as a combination of standard loss functions.
Specifically, the compression term is expressed using the kl-divergence as described in \Cref{eq: kl_div}. 
Here, $r(A)$ and $r(Z)$ are treated as fixed spherical Gaussians, represented as $\mathcal{N}(0,I)$.
The sufficiency term uses the reconstruction loss between input and $q_{\phi}(X|A,Z)$-generated samples, while the prediction term employs the CrossEntropy loss.

\subsection{Adversarial Robustness and Generalization}
In this section, we elucidate how our DisTIB finds a pair of disentangled representations $(A,Z)$ to enhance model generalization.
In our implementation, the representation $A$ solely encompasses the label-related information imperative for prediction, sidelining the remaining sample-specific details.
Consequently, akin to \cite{alemi2016deep,pan2021disentangled}, we assess the model's generalization capability by evaluating the accuracy on the test set using the learned label-related representation $A$. 
Additionally, \cite{szegedy2013intriguing,goodfellow2014explaining} highlighted that deterministic models trained with maximum likelihood estimation are susceptible to adversarial samples, \ie, input with slight perturbations that retain the appearance of natural images but are deliberately designed to deceive the model.
Building on this understanding, studies such as \cite{tsipras2018there,hosseini2019dropping} emphasized the potential benefits of limiting information in representations to bolster adversarial robustness. Aligned with this perspective, we further evaluate our model's adversarial robustness by focusing on the learned label-related representation $A$.

\subsubsection{Datasets}
Following previous works \cite{alemi2016deep,pan2021disentangled}, we assess DisTIB using established benchmarks for disentangled representation learning: MNIST, FashionMNIST, and CIFAR10. For MNIST and FashionMNIST, we adhere to the standard split, using 60000 samples for training and 10000 for testing. For CIFAR10, we employ a 50000/10000 split for train/test samples, respectively.

\subsubsection{Experiment Setting}
In this section, we delineate the aforementioned evaluation metrics, \ie, generalization and adversarial robustness, crucial for understanding the efficacy of our DisTIB.
Firstly, in line with \cite{alemi2016deep,pan2021disentangled}, we gauge generalization by computing the average accuracy on the test set after training on train set.
Secondly, we employ the same attack method \cite{goodfellow2014explaining} as in \cite{alemi2016deep,pan2021disentangled} for a fair comparison.
This attack strategy synthesizes adversarial samples by taking a single gradient step, where the perturbation magnitude for each pixel is regulated by the hyperparameter $\epsilon$. 
After training the model, we begin with adversarial sample generation on both train and test sets.
We then evaluate the model's adversarial robustness on generated adversarial samples by measuring the mean classification accuracy.

\subsubsection{Results}
Following \cite{pan2021disentangled}, we evaluate DisTIB against prominent IB-based disentangled representation learning methods, including regularization-based methods \cite{alemi2016deep,rodriguez2020convex,kolchinsky2019nonlinear,yang2023snib,wu2023exponential} and disentanglement-based methods \cite{pan2021disentangled,dang2024disentangled}. 
\Cref{table_generalization_attack} summarizes our experimental results on both generalization and adversarial robustness.
In terms of generalization, the proposed DisTIB showcases superior generalization ability, outperforming previous state-of-the-art (SOTA) methods by around 1\% on average.
When evaluating adversarial robustness, our DisTIB consistently achieves the best performance under different perturbation magnitudes across all three datasets. 
Specifically, taking MNIST as an instance, our DisTIB delivers an impressive 94.8\% accuracy under $\epsilon=0.1$, whereas other competitors fall short, with the closest competitor, DisGenIB, achieving an accuracy of 90.7\%.
Additionally, although previous regularization based methods explore various mutual information constraints, \eg, combinational \cite{yang2023snib} or nonlinear \cite{wu2023exponential}, for information compression, they inevitably suffer from label-related information loss as the level of compression intensifies \cite{pan2021disentangled}, \ie, the notorious performance-compression trade-off.
	As a result, such trade-off leads to prediction performance degeneration and weakens their adversarial robustness \cite{pan2021disentangled}.
In contrast, when comparing with disentanglement based methods, although DisenIB and DisGenIB demonstrate commendable results in adversarial robustness with elaborately designed disentanglement objectives, they still lag behind our DisTIB by more than 3\% in classification accuracy.
We attribute this performance disparity to the adversarial training in DisenIB and the objective factorization in DisGenIB, \ie, complicated optimization, leading to suboptimal disentanglement and thus limiting the adversarial robustness performance.
As a result, these empirical results underscore the appealing generalization and adversarial robustness of our DisTIB, thereby demonstrating its efficacy in disentangled representation learning.

\begin{figure*}[t]
  \centering
  \subfloat[MNIST]{\includegraphics[width=0.3\textwidth,height=0.3\textwidth]{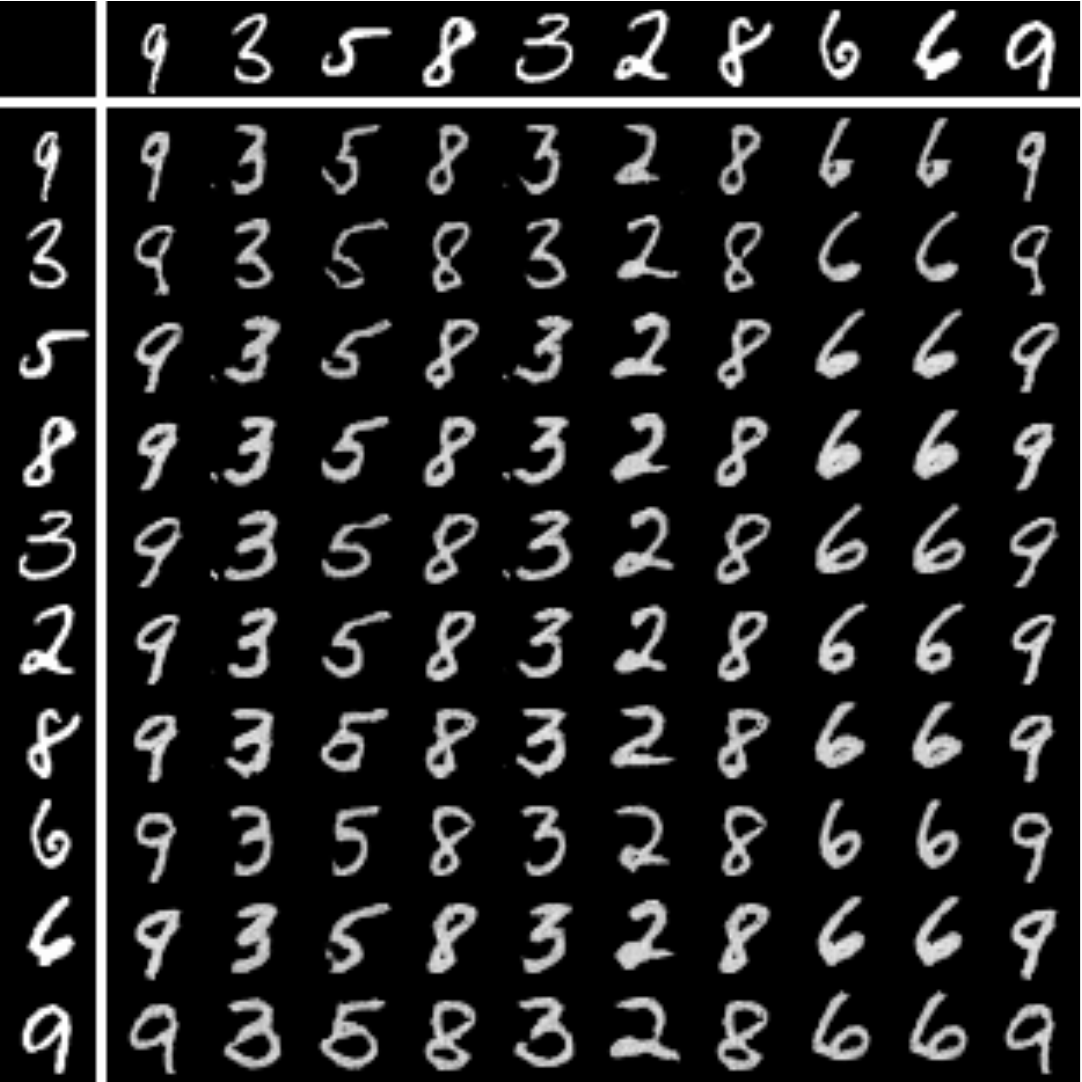}\label{fig: MNIST}}\hfill
  \subfloat[Sprites]{\includegraphics[width=0.3\textwidth,height=0.3\textwidth]{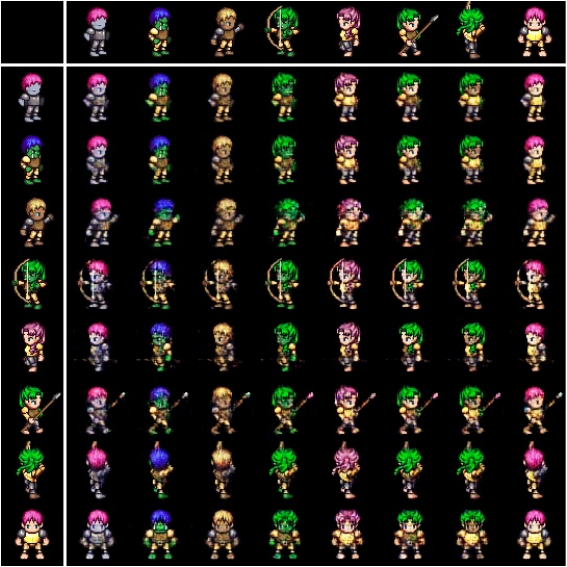}}\hfill
  \subfloat[dSprites]{\includegraphics[width=0.3\textwidth,height=0.3\textwidth]{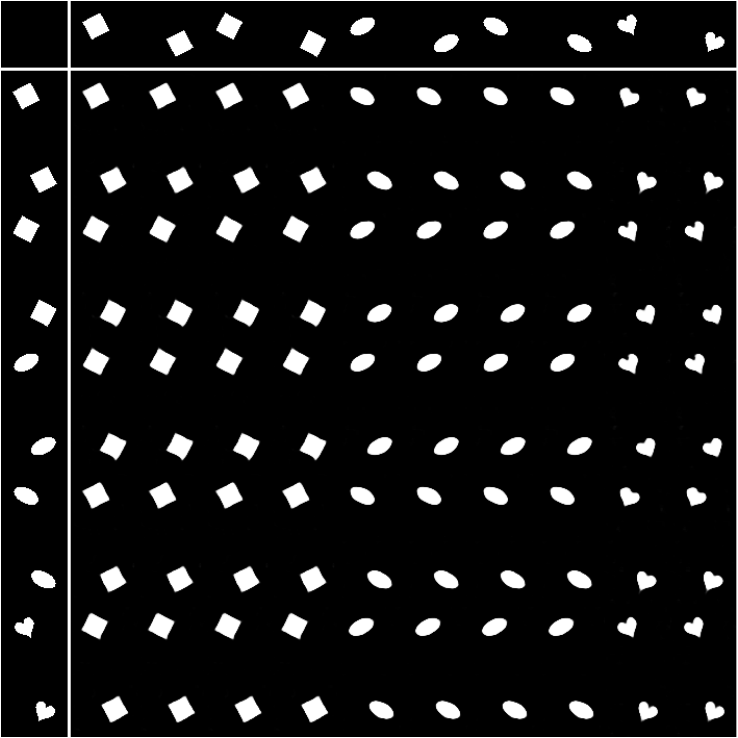}}
  \caption{Visualization of disentangled sample generation, where samples are generated by the label-related information $A$ of the top row and sample-exclusive information $Z$ of the leftmost column. 
The top row and leftmost column images come from the dataset and the diagonal images show reconstructions.
}\label{fig: disentangle}
\end{figure*}

\subsection{Supervised Disentangling}
In addition to quantitative evaluations of DisTIB's disentanglement behavior, we further explore supervised disentangling to intuitively demonstrate independent feature manipulation, serving as qualitative benchmarks.
Specifically, from a dataset with annotated labels, we first employ specific encoders to obtain disentangled representation pairs from the input. Subsequently, we employ the generator to synthesize samples based on these representations.
This visualization provides an intuitive insight into both the label-related and sample-exclusive information captured by our DisTIB.
\subsubsection{Datasets}
Following \cite{mathieu2016disentangling,pan2021disentangled}, we qualitatively evaluate DisTIB's disentanglement efficacy on widely used benchmark datasets: MNIST \cite{deng2012mnist}, Sprites \cite{reed2015deep} and dSprites \cite{dsprites17}.
\begin{itemize}
  \item \textbf{MNIST:} This dataset is renowned for its grayscale images of handwritten digits, ranging from 0 to 9. In this dataset, the label-related information refers simply to the class of digit content, while the sample-specific information denotes the writing style.
  \item \textbf{Sprites:} This dataset is composed of 672 characters from 20 different animations. Each sprite image can be characterized by seven distinct attributes: body, gender, hair, armor, arm, greaves, and weapon. In our implementation, we use gender, hair, armor and greave as label-related information, while the remaining attribute as sample-exclusive information.
  \item \textbf{dSprites:} This dataset encompasses three fundamental 2D shapes: squares, circles, and hearts. These shapes are systematically varied across different rotations and positions within each image. For our study, the inherent 2D shape is treated as label-related information. In contrast, variations in rotation and position are considered as sample-exclusive information.
\end{itemize}

\subsubsection{Experiment Setting}
The goal of supervised disentangling is to show that our DisTIB can effectively disentangle the sample-exclusive information from the label-related information of the input and synthesize satisfactory analogies.
Following previous works \cite{mathieu2016disentangling,pan2021disentangled,hadad2018two,kim2018disentangling}, we adopt a qualitative evaluation approach using the ``swapping" paradigm.
Specifically, given an image pair $(I_1, I_2)$, we leverage specific encoders $p(A|X),p(Z|X)$ to extract disentangled representation pairs $(A_1,Z_1)$ and $(A_2,Z_2)$, respectively.
Subsequently, we use generator $q_\phi$ to synthesize samples conditioning on the label-related information extracted from one image and sample-exclusive information obtained from the other, resulting in combinations $(A_1,Z_2)$ and $(A_2,Z_1)$.
Given this framework, the ideal ``swapping'' generation should seamlessly incorporate label-related and sample-exclusive information from corresponding original images.
\subsubsection{Results}
\Cref{fig: disentangle} shows the samples generated in ``swapping'' setting across all three datasets.
Overall, DisTIB generates faithful samples using various disentangled representation pairs $(A,Z)$.
Taking the dSprites as an example, a column-wise observation reveals that the label-related information (shape) from the topmost row, whether square, circle, or heart-shaped, is consistently retained from the original images. 
On the other hand, a row-wise perspective shows that the sample-specific information (location and rotation) from the leftmost column is seamlessly integrated into the synthesized samples.
Significantly, although the label-related and sample-specific information originates from different samples, their integration is seamlessly achieved in the generated images. This attests to DisTIB's optimal disentanglement, ensuring no overlap or loss of either type of information.
Additionally, experiments on the other two datasets show consistent trends: by retaining the label-related information from one image and imposing the sample-specific information from another, the synthesized images display a coherent blend of these attributes. 
These results highlight DisTIB's efficacy in disentangling label-related and sample-specific information.

\begin{table*}[!t]
      \centering
      \caption{Comparison with SOTA methods on miniImageNet and tieredImageNet. RL and SG denote representation learning and sample generation, respectively. The top two results are shown in bold and underlined.}
        \label{tab:mini_tiered_results}
        \begin{tabular}{lccccc}
        \toprule 
        \multicolumn{1}{l}{\multirow{2}{*}{\textbf{Method}}}
        & \multicolumn{1}{c}{\multirow{2}{*}{\textbf{Backbone}}} 
        & \multicolumn{2}{c}{\textbf{miniImageNet}} 
        & \multicolumn{2}{c}{\textbf{tieredImageNet}} 
        \\ 
        \cmidrule(l){3-6}
        \multicolumn{1}{c}{}
        & \multicolumn{1}{c}{}   
        & \textbf{$5$-way $1$-shot} & \textbf{$5$-way $5$-shot}  & \textbf{$5$-way $1$-shot}  & \textbf{$5$-way $5$-shot} \\  
        \midrule
        \midrule
        CGCS  \cite{gao2021curvature} (ICCV'21)                                           & BigResNet-12                           & $71.79 \pm 0.23\%$               & $83.00 \pm 0.17\%$             & \underline{$77.19 \pm 0.24\%$}                 & $86.18 \pm 0.15\%$                 \\
        RENet  \cite{kang2021relational}  (ICCV'21)                                                & ResNet-12                           & $67.60 \pm 0.44\%$               & $82.58 \pm 0.30\%$             & $71.61 \pm 0.51\%$                 & $85.28 \pm 0.35\%$                 \\
        PAL  \cite{ma2021partner}   (ICCV'21)                                                & ResNet-12                          & $69.37 \pm 0.64\%$              & $84.40 \pm 0.44\%$             & $72.25 \pm 0.72\%$                 & $86.95 \pm 0.47\%$                 \\
        InvEq  \cite{rizve2021exploring}  (CVPR'21)                                                & ResNet-12                           & $67.28 \pm 0.80\%$              & $84.78 \pm 0.52\%$             & $71.87 \pm 0.89\%$                 & $86.82 \pm 0.58\%$                 \\
        ODE  \cite{xu2021learning}   (CVPR'21)                                       & ResNet-12                          & $67.76 \pm 0.46\%$              & $82.71 \pm 0.31\%$             & $71.89 \pm 0.52\%$                 & $85.96 \pm 0.35\%$                 \\
        FRN  \cite{wertheimer2021few}  (CVPR'21)                             & ResNet-12                     & $66.45 \pm 0.19\%$              &  $82.83 \pm 0.13\%$            & $72.06 \pm 0.22\%$                 & $86.89\pm0.14\%$                 \\
        DeepEMD \cite{zhang2022deepemd}  (TPAMI'22)                                               & ResNet-12                          & $68.77 \pm 0.29\%$              & $84.13 \pm 0.53\%$             & $74.29 \pm 0.32\%$           & $86.98 \pm 0.60\%$        \\
        FeLMi \cite{roy2022felmi} (NIPS'22)       & ResNet-12                           & $67.47\pm0.78\%$              & \underline{$86.08\pm0.44\%$}             & $71.63\pm0.89\%$                 & $87.07\pm0.55\%$                 \\
        META-NTK \cite{wang2022global}  (CVPR'22)                                          & ResNet-12                          & $64.26 \pm 0.14\%$              & $80.35 \pm 0.12\%$            & $72.37 \pm 0.79\%$                 & $86.43 \pm 0.53\%$                 \\
        MCL \cite{liu2022learning}   (CVPR'22)      & ResNet-12                           & $69.31\%$              & $85.11\%$             & $73.62\%$                 & $86.29\%$                 \\
        SVAE  \cite{xu2022generating}   (CVPR'22)                                                 & ResNet-12                          &  \underline{$74.84 \pm 0.23\%$}             & $83.28 \pm 0.40\%$             & $76.98 \pm 0.65\%$                 & $85.77 \pm 0.50\%$                 \\
        DeepBDC \cite{xie2022joint}   (CVPR'22)                                               & ResNet-12                          & $67.34 \pm 0.43\%$              & $84.46 \pm 0.28\%$             & $72.34 \pm 0.49\%$                 & \underline{$87.31 \pm 0.32\%$}        \\
        CC \cite{yang2022few}   (ECCV'22)                                           & ResNet-12                           & $70.19 \pm 0.46\%$              & $84.66\pm 0.29\%$             & $72.62 \pm 0.51\%$                 & $86.62\pm 0.33\%$ \\
        tSF \cite{lai2022tsf}  (ECCV'22)                              & ResNet-12                          & $69.74 \pm 0.47\%$              & $83.91 \pm 0.30\%$             & $71.98 \pm 0.50\%$                 & $85.49 \pm 0.35\%$       \\
        HTS \cite{zhang2022tree}   (ECCV'22)                                             & ResNet-12                          & $64.95 \pm 0.18\%$              & $83.89 \pm 0.15\%$             & $68.38 \pm 0.23\%$                 & $86.34 \pm 0.18\%$        \\
        HGNN \cite{chen2021hierarchical}  (TCSVT'22)                                        & ResNet-12                          & $60.03 \pm 0.51\%$              & $79.64 \pm 0.36\%$             & $64.32 \pm 0.49\%$                 & $83.34 \pm 0.45\%$                 \\
        TEDC \cite{zhang2022task}  (TCSVT'22)                               & ResNet-12                          & $67.31 \pm 0.20\%$              & $83.22 \pm 0.16\%$             & $72.29 \pm 0.22\%$                 & $86.22 \pm 0.16\%$                 \\
        OSLO \cite{Boudiaf_2023_CVPR}    (CVPR'23)                                               & ResNet-12                          & $71.73 \pm 0.72\%$              & $83.40 \pm 0.44\%$            & $76.64 \pm 0.74\%$                 & $86.35 \pm 0.52\%$                \\
        FGFL \cite{Cheng_2023_ICCV}       (ICCV'23)                                            & ResNet-12                          & $69.14 \pm 0.80\%$              & $86.01 \pm 0.62\%$            & $73.21 \pm 0.88\%$                 & $87.21 \pm 0.61\%$                \\
        \midrule
        			  DisTIB + RL                                                    & ResNet-12                          & 69.34 $\pm$ {0.73}\%              & {84.79} $\pm$ {0.52}\%            & {{73.05} $\pm$ {0.91}}\%                 & {{86.91} $\pm$ {0.59}}\%                \\
        DisTIB + SG                                                     & ResNet-12                          & \textbf{77.19} $\pm$ \textbf{0.88}\%              & \textbf{86.43} $\pm$ \textbf{0.52}\%             & \textbf{77.60} $\pm$ \textbf{0.89}\%                 & \textbf{87.35} $\pm$ \textbf{0.58}\%\\
        \bottomrule
        \end{tabular}
        \end{table*}

\begin{figure}[!t]
		\centering
		\subfloat[Hair $\rightarrow$ Bald.]{{\includegraphics[width=0.95\linewidth]{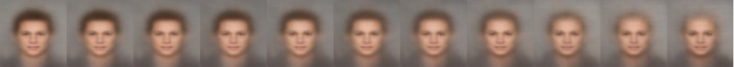}}}\\
		\subfloat[Dark Skin $\rightarrow$ Pale Skin.]{{\includegraphics[width=0.95\linewidth]{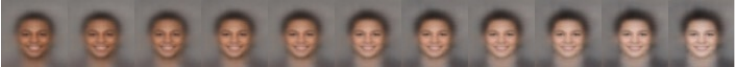}}}\\
		\subfloat[Unsmile $\rightarrow$ Smile.]{{\includegraphics[width=0.95\linewidth]{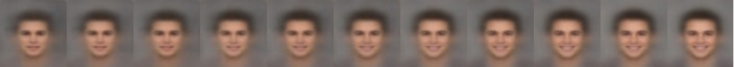}}}\\
		\subfloat[Female $\rightarrow$ Male; No Beard $\rightarrow$ Mustache; Makeup $\rightarrow$ No Makeup.]{{\includegraphics[width=0.95\linewidth]{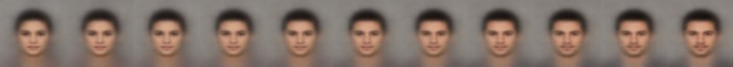}}\label{fig: multi attr}}\\
		\caption{
  Illustration of applying our DisTIB to disentangle facial attributes.
		Each row displays the generation results of a specific facial attribute interpolation, where the leftmost/rightmost image has an attribute value of 0/1; intermediate images represent interpolated attributes with a step size of 0.1.
  }
		\label{fig: CelebA}
	\end{figure}

\subsection{Facial Attribute Disentangling}
We further apply DisTIB to the more challenging and realistic CelebA \cite{liu2015deep} dataset to qualitatively analyze its disentanglement performance in complex real-world data. Specifically, we focus on disentangling facial attributes in the CelebA dataset to enable controllable facial generation. The specific experimental details are as follows:

	\subsubsection{Implementation Details}
	We evaluate DisTIB on CelebA \cite{liu2015deep}, a standard facial image benchmark containing 202,599 images of 10,177 identities. Each image is annotated with 40 binary attributes that describe facial features, such as smiling, wearing glasses, and hairstyle. Following previous work \cite{ yang2021l2m}, we use attribute interpolation to generate new images for showcasing our DisTIB's disentanglement performance. Specifically, given a facial image, we select an attribute to modify and perform interpolation, gradually changing the attribute value with a step size of 0.1 while keeping others constant, resulting in 10 interpolated images.

	\subsubsection{Results}
	\Cref{fig: CelebA} showcases the superior performance of DisTIB on the CelebA dataset by effectively disentangling different facial attributes to control the appearance of generated images, including transitions such as from having hair to being bald, from black hair to blonde hair, and from not wearing glasses to wearing glasses.
	Furthermore, as shown in \Cref{fig: multi attr}, even in multi-attribute interpolation experiments, our images exhibit smooth and natural transitions of facial features while maintaining facial consistency. This highlights that the representations learned by DisTIB effectively capture attribute-related information while ensuring disentanglement among attributes.
	These results further demonstrate the efficacy of DisTIB when handling complex real-world data, highlighting its appealing applicability.

\subsection{Few-Shot Learning (FSL)}
In this section, beyond traditional disentangled representation learning benchmarks, we further extend DisTIB to the more challenging downstream task of few-shot learning. 
Specifically, FSL focuses on mining label-related information from a limited number of labeled samples to match with the unlabeled samples for classification \cite{liu2022intermediate}. However, due to the data scarcity inherent in FSL, its performance often suffers from spurious correlations in the learned representations \cite{dang2023counterfactual,yue2020interventional}. 
To this issue, we aim to utilize DisTIB to disentangle label-related information from sample-specific information, thereby facilitating more effective knowledge transfer for FSL.

\subsubsection{Datasets}
We evaluate DisTIB on standard FSL benchmarks: miniImageNet \cite{vinyals2016matching} and tieredImageNet \cite{renmeta}.
MiniImageNet and tieredImageNet comprise 100 classes with 600 (84$\times$84) images each and 608 classes with about 1200 (84$\times$84) images each, respectively. 
Following \cite{vinyals2016matching}, miniImageNet is randomly divided into 64, 16, 20 classes for train, valid and test, respectively.
In contrast, tieredImageNet organizes its classes into 34 high-level categories with divisions of 20, 6, and 8 for train, valid, and test, respectively.

In the test stage, FSL tasks are given in the formulation of N-way K-shot, where N classes are sampled with K labeled samples per class, denoted as support set $\mathcal{S}$. Meanwhile, a query set $\mathcal{Q}$ with sufficient samples is sampled from the same support categories for FSL evaluation.
Finally, following previous work \cite{snell2017prototypical,rizve2021exploring,zhang2022deepemd}, we report results on 600 randomly sampled standard 5-way 1/5-shot FSL tasks in terms of mean accuracy and 95\% confidence interval.
\subsubsection{Methods} 
Inspired by prior works, we utilize the disentangled representations $(A,Z)$ learned by DisTIB in the following manner:
\textbf{1)}
Following \cite{rizve2021exploring}, we train a classifier based on the label-related information $A$ and corresponding labels, denoted as the \textbf{representation learning (RL)} method.
\textbf{2)} Inspired by FeLMi \cite{roy2022felmi} and SVAE \cite{xu2022generating}, we tap into DisTIB's capability for disentangled generation to synthesize both discriminative and diverse samples, thereby alleviating the data scarcity in FSL.
Specifically, as in \Cref{alg: inference}, we combine various $A$ and $Z$ to synthesize additional samples, whose label is assigned by its corresponding label-related information $A$.
Subsequently, as in ProtoNet, we average sample features within classes on both support and synthesized samples to derive the \textbf{sample generation (SG) prototype}.
To further eliminate the bias in the prototype, the SG prototype undergoes rectification based on the strategy proposed in \cite{zhang2021prototype}.
Query samples are then matched using the nearest neighbor classifier to the refined prototype.

\begin{table}[!t]
		\centering
		\caption{Fine-grained classification with few-shot setting results on the CUB dataset, where all results are obtained with ResNet-12 backbone.}
		\begin{tabular}{l c c}
			\toprule
			\multirow{2}{*}{Method}  & \multicolumn{2}{c}{CUB} \\
			\cmidrule(lr){2-3}
				&    1-shot & 5-shot \\
			\midrule
			FEAT \cite{ye2020few}   & 75.68$\pm$0.20 & 87.91$\pm$0.13 \\
			BML \cite{zhou2021binocular}  & 76.21$\pm$0.63 & 90.45$\pm$0.36 \\
			DeepEMD \cite{zhang2022deepemd}   & 79.27$\pm$0.29 & 89.80$\pm$0.51 \\
			ProtoNet \cite{snell2017prototypical}  & 72.25$\pm$0.21 & 87.47$\pm$0.13 \\
			DFR \cite{cheng2023disentangled}   & 78.07$\pm$0.79 & 89.74$\pm$0.51 \\
			VFD \cite{xu2021variational}   & 79.12$\pm$0.83 & 91.48$\pm$0.39 \\
			FGFL \cite{Cheng_2023_ICCV}  & 80.77$\pm$0.90 & 92.01$\pm$0.71 \\
			TEDC \cite{zhang2022task}  & 76.11$\pm$0.21 & 89.54$\pm$0.12 \\
			\midrule
			Ours  & \textbf{82.03$\pm$0.51} & \textbf{92.24$\pm$0.39} \\
			\bottomrule
		\end{tabular}
		\label{tab:cub_results}
	\end{table}

\subsubsection{Results}
For a fair comparison, we report results of standard FSL competitors in \Cref{tab:mini_tiered_results}, excluding influences such as priors, enhanced backbones, high-resolution inputs, \etc 
 Specifically, even though previous representation learning-based methods have explored various strategies, \eg, self-supervised learning \cite{rizve2021exploring} and distribution calibration \cite{zhang2022task}, to enhance the robustness of representations, they still suffer from the entanglement between label-related information and sample-exclusive information.
In contrast, ``DisTIB + RL'' effectively disentangle label-related information with certifiable efficacy, thereby highlighting a 2\% performance improvement on average.
Furthermore, ``DisTIB + SG" is able to synthesize numerous discriminative and diverse extra samples by further utilizing the sample-exclusive information.
Notably, it surpasses previous data augmentation based methods \cite{xu2022generating,roy2022felmi} by roughly 1\%, underscoring the exceptional quality of samples synthesized by DisTIB's disentangled generation.
As a result, by adapting DisTIB to few-shot learning, we validate its efficacy in disentangled representation learning and underscore the importance of such representations for downstream tasks.

\begin{figure}[t]
    \centering
    \subfloat[train set.]{\includegraphics[width=0.5\linewidth]{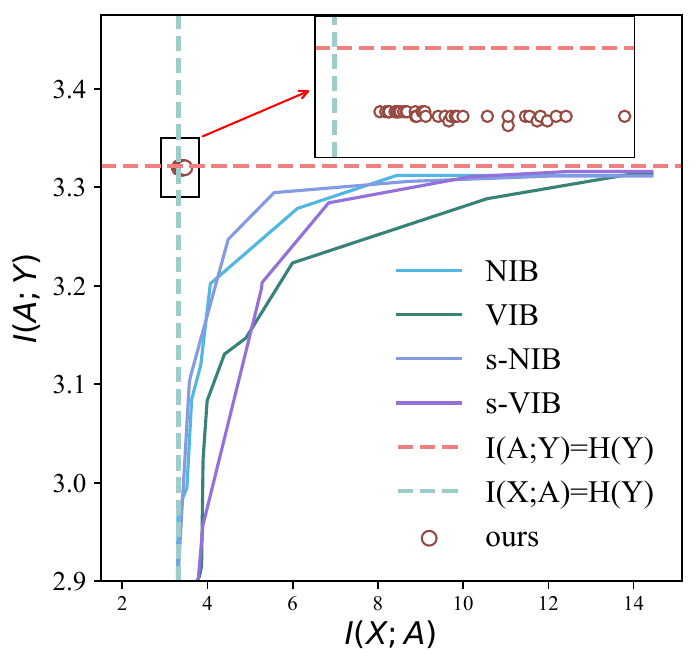}}\hfill
    \subfloat[test set.]{\includegraphics[width=0.5\linewidth]{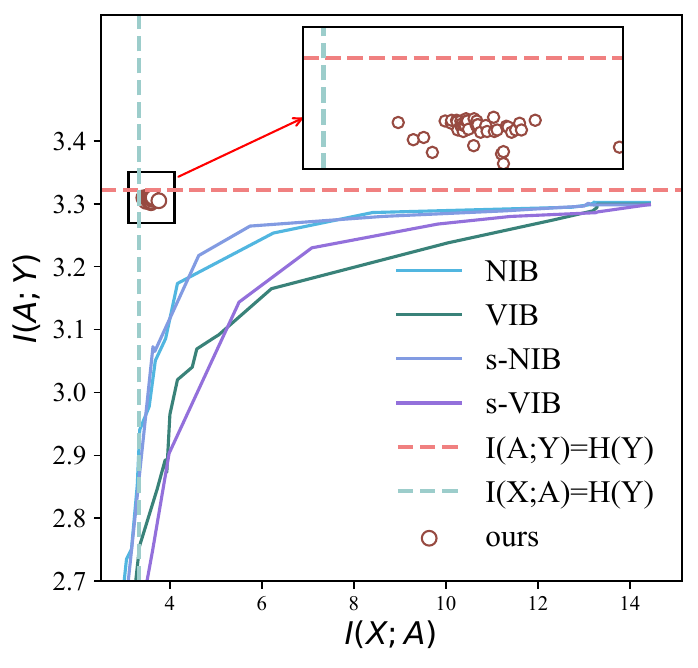}}
    \caption{The performance-compression trade-off on MNIST, where DisTIB is represented as dots due to its convergence to optimal disentanglement.}
    \label{fig: tradeoff}
\end{figure}

\begin{figure*}[!t]
    \centering
    \subfloat[$I(X;A)$ train.]{\includegraphics[width=0.25\textwidth]{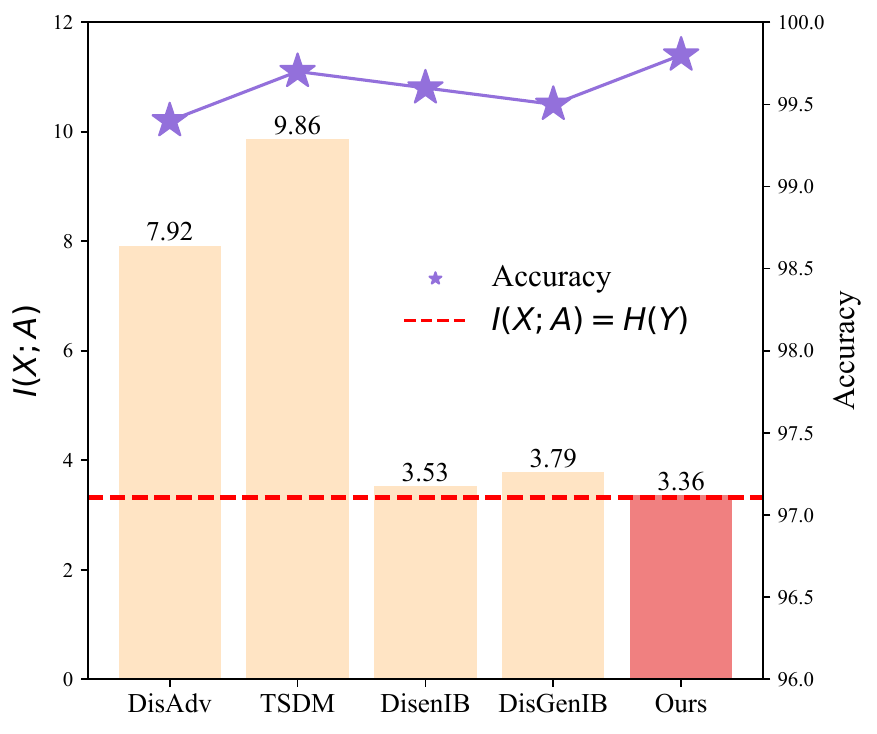}}\hfill
    \subfloat[$I(X;A)$ test.]{\includegraphics[width=0.25\textwidth]{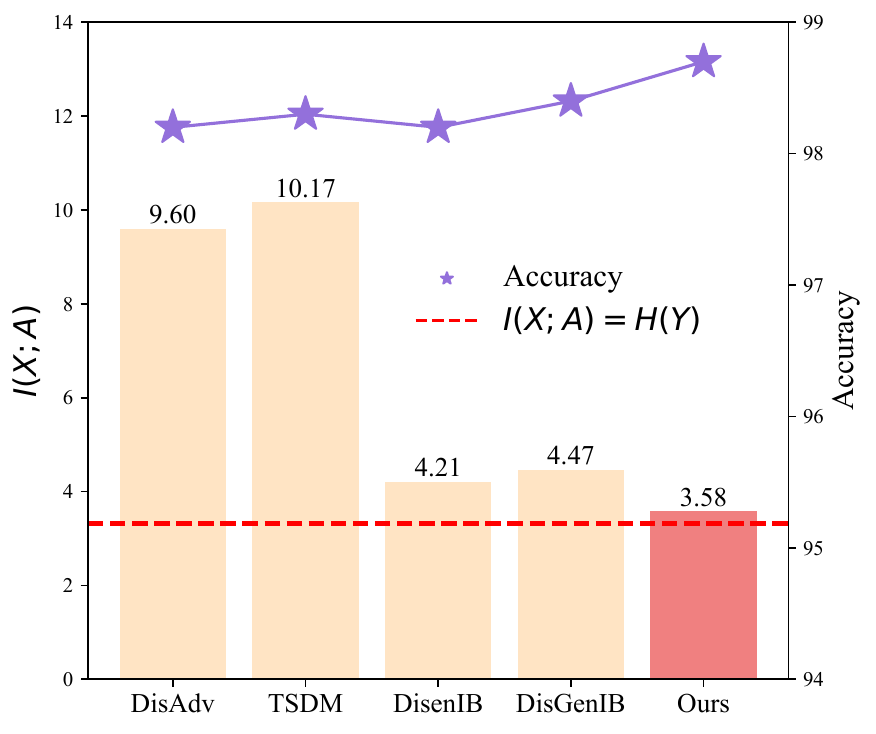}}\hfill
    \subfloat[$I(X;Z)$ train.]{\includegraphics[width=0.25\textwidth]{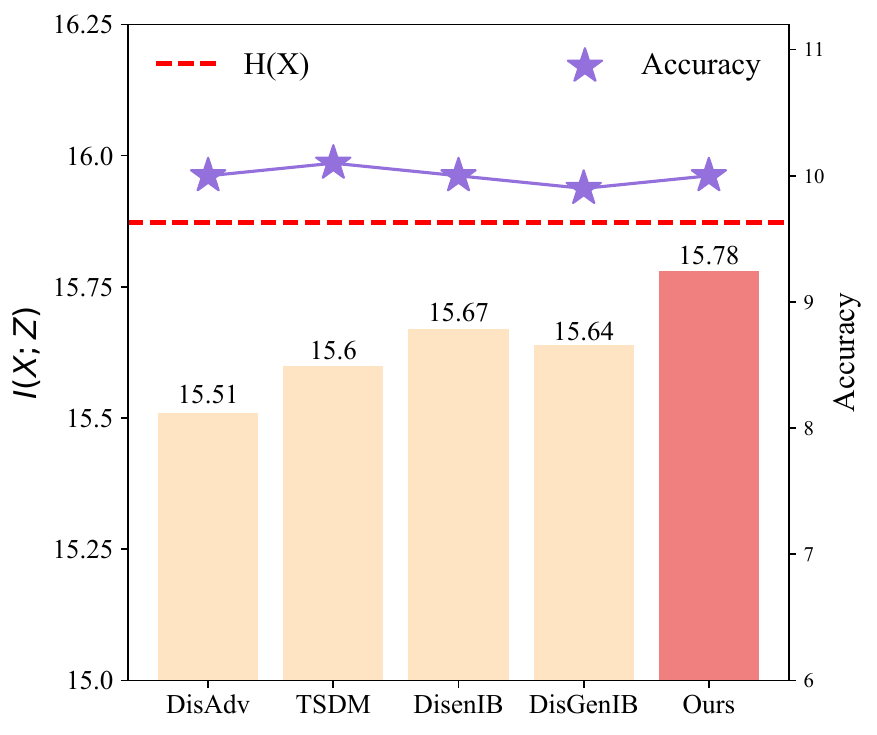}}\hfill
    \subfloat[$I(X;Z)$ test.]{\includegraphics[width=0.25\textwidth]{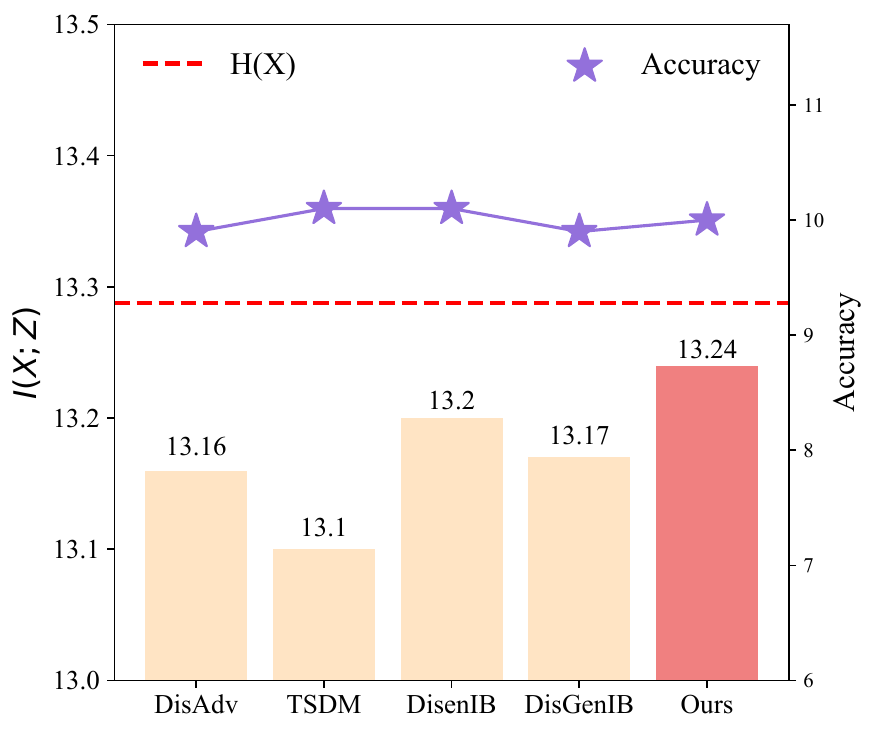}}
    \caption{MNIST experiments results of mutual information quantization and classification accuracy on disentangled representations, where the red line denotes the ideal objective. Note that $I(X;Z)=H(X)-H(X|Z)$, where the second term in RHS has no analytical solution, thus we report $H(X)$ as an upper bound of $I(X;Z)$ in (c) and (d).}\label{fig: MI quant}
\end{figure*}

\begin{figure*}[!t]
    \centering
    \subfloat[\scriptsize Entangled Sprites.]{\includegraphics[width=0.24\textwidth,height=0.22443142\textwidth]{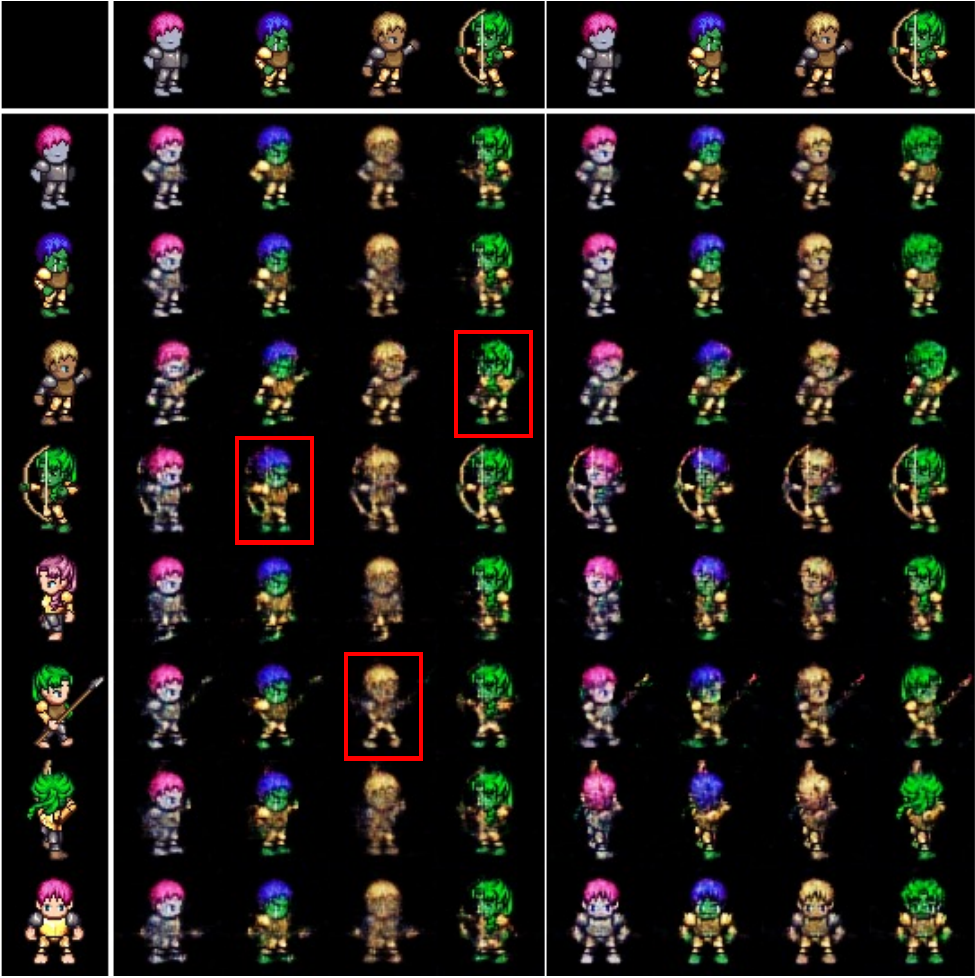}\label{fig: sprites_bad}}\hfill
    \subfloat[\scriptsize Entangled sample generation.]{\includegraphics[width=0.24\textwidth]{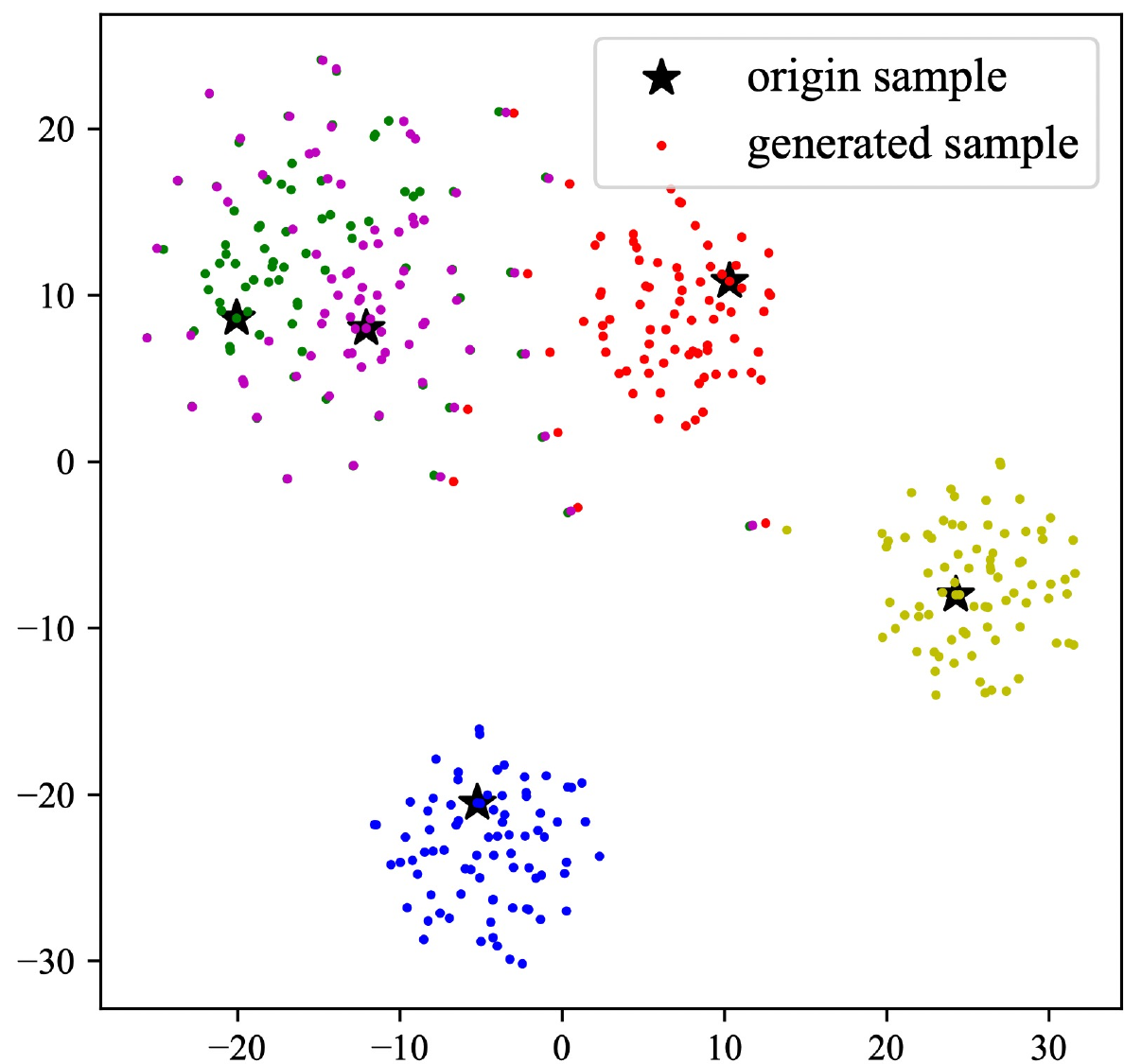}\label{fig: gen_bad}}\hfill
    \subfloat[\scriptsize Disentangled sample generation.]{\includegraphics[width=0.24\textwidth]{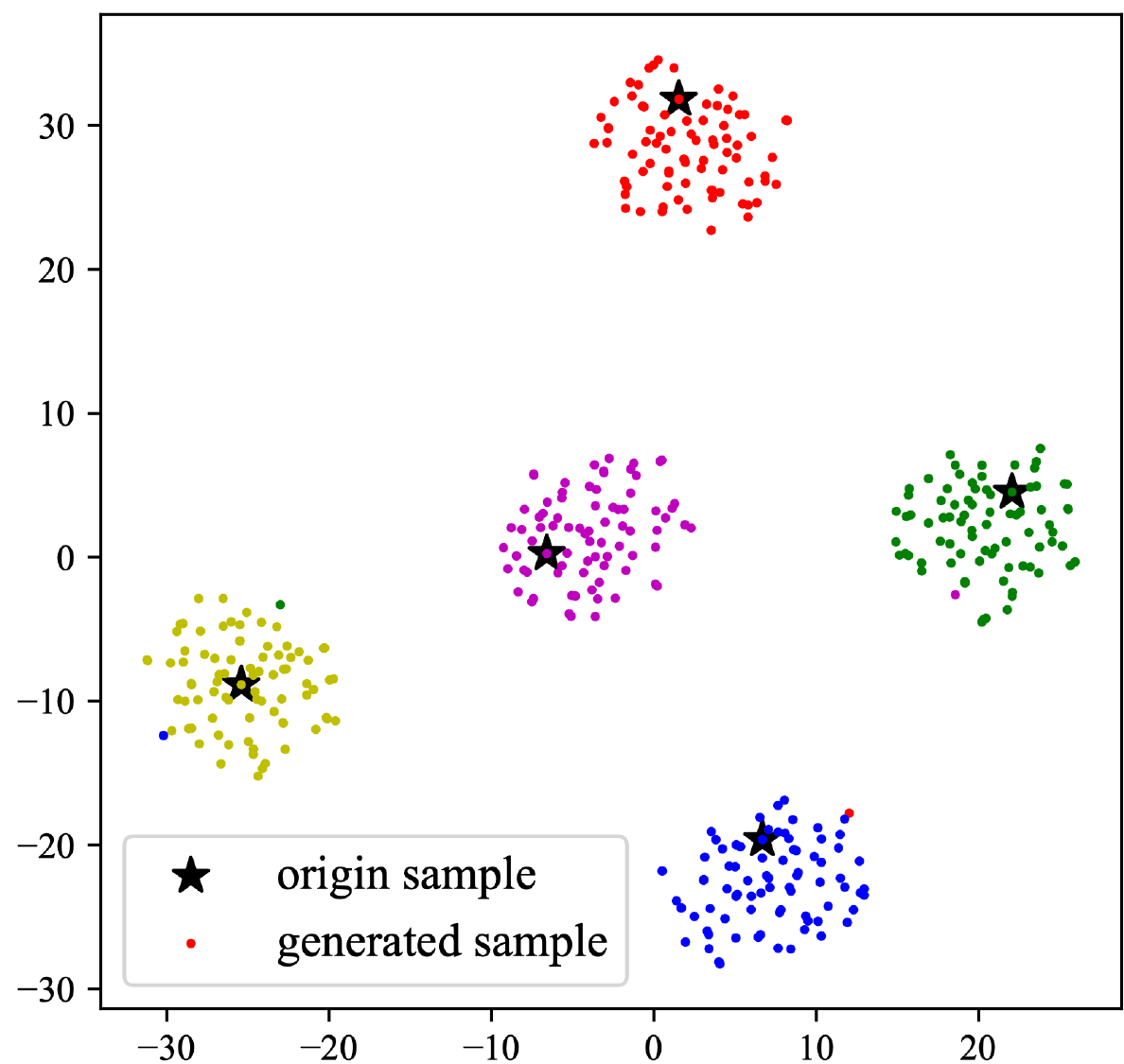}\label{fig: gen_good}}\hfill
    \subfloat[\scriptsize Disentangled prototypes.]{\includegraphics[width=0.24\textwidth]{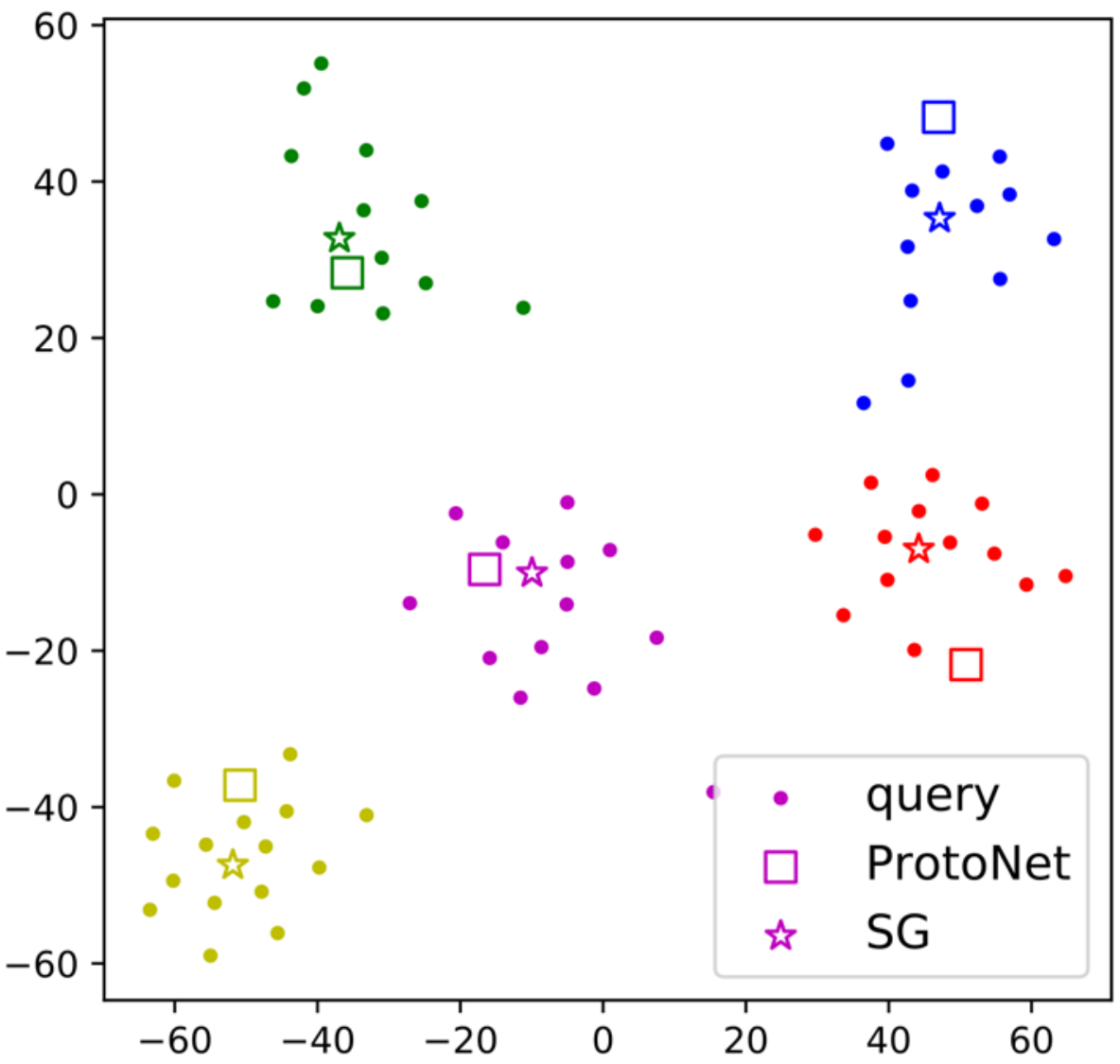}\label{fig: proto_vis}}
    \caption{The left/right part in (a) are Sprites synthesized by DisTIB without/with disentanglement.
    (b) and (c) are FSL sample generations of DisTIB without or with disentanglement. (d) is the visualization of prototype and query samples, where SG is prototype estimated by our sample generation.}
\end{figure*}
\subsection{Fine-Grained Classification}
In this section, to better showcase the efficacy of our DisTIB, we apply it to the challenging task of fine-grained classification, which involves samples from different classes with highly similar visual features and minimal inter-class differences.
Therefore, this task requires representations to capture the subtle differences between categories while reducing feature redundancy and confusion.
To address this issue, we employ DisTIB to encode only label-related information while discarding sample-specific information, thereby better handling the fine-grained inter-class differences.

\subsubsection{Implementation Details}
We evaluate DisTIB on standard fine-grained classification benchmark with few-shot setting: CUB \cite{wah2011caltech}.
Specifically, the CUB dataset consists of 11,788 images spanning 200 bird species. Following \cite{xu2021variational}, we divided the dataset into 100 classes for training, 50 classes for validation, and the remaining 50 classes for testing. 
Moreover, following ProtoNet \cite{snell2017prototypical}, we employ the first-order statics (mean) of label-related information $A$ learned by our DisTIB as the class prototype, and the evaluation implementation and metrics follow the same protocols used in the FSL setting.

\subsubsection{Results}
As summarized in \Cref{tab:cub_results}, our DisTIB showcases consistent performance superiority on the CUB dataset. 
Notably, the strong competitor VFD \cite{xu2021variational} simply extracts label-related features by average pooling the output of the feature extractor for subsequent disentanglement.
In contrast, our DisTIB benefits from an elaborately designed disentanglement objective, which enables DisTIB to extract precise label-related information and effectively capture subtle inter-class differences.
 Thus, DisTIB improves performance by an average of 2\%, highlighting the potential and necessity of representation disentanglement in fine-grained tasks.

\subsection{Discussion and Ablation Study}
In this section, we first conduct experiments on the information plane to validate our theoretical analyses (\textbf{RQ2}). Subsequently, we ablate different components of DisTIB to elucidate how it facilitates disentanglement (\textbf{RQ3}).
\subsubsection{Behavior on Information Plane}
To study DisTIB and previous regularization-based competitors \cite{alemi2016deep,kolchinsky2019nonlinear,rodriguez2020convex} in detail, we conduct experiments to explore their performance-compression tradeoff by tuning $\beta$. These results form a curve on the information plane shown in \Cref{fig: tradeoff}, defined by the $I(X;A)$ (x-axis, compression) and $I(A;Y)$ (y-axis, performance). 
Although searching the above trade-off can help previous methods eliminate part of redundant information, their performance drops when further compressing the information to $I(X;A)=H(Y)$, which fails to achieve optimal disentanglement.
In contrast, DisTIB is robust to $\beta$ and closest to the optimal disentanglement,
indicating it can effectively disentangle the label-related information. 
These results qualitatively demonstrate the best performance-compression trade-off and disentanglement performance of DisTIB, and the validity of our theoretical analyses.

\begin{table}[t]
  \tabcolsep = 4 pt
      \centering
        \caption{Influence of regularizers on MNIST adversarial robustness and miniImageNet FSL. }
      \label{tab: ablation loss}
      \begin{tabular}{cc|cc|cc}
      \toprule 
      \multicolumn{2}{c|}{Regularizer} &
      \multicolumn{2}{c|}{\emph{Adv} Robustness} &
      \multicolumn{2}{c}{FSL (1-shot)}
      \\
      \multicolumn{1}{c}{\emph{Disen}}
      & \multicolumn{1}{c|}{Sufficiency}
      & \multicolumn{1}{c}{$\epsilon=0.1$} 
      & \multicolumn{1}{c|}{$\epsilon=0.3$} 
      & \multicolumn{1}{c}{$RL$} 
      & \multicolumn{1}{c}{$SG$} 
      \\ 
      
      \midrule
            &            & $88.24$  &$9.63$              & $65.94$  & N/A\\
       \checkmark       & & $92.67$  &$69.92$            & $68.07$  & N/A\\
            & \checkmark & $85.44$  &$6.53$            & $64.82$  &$69.27$\\
       \checkmark       & \checkmark & \textbf{94.76}  &\textbf{74.38}  & \textbf{69.34}  &\textbf{77.19}\\
      \bottomrule
      \end{tabular}
      \end{table}

\subsubsection{Comparison With Explicit Disentanglement Methods}
To substantiate the efficacy of DisTIB's implicit disentanglement, we conduct experiments to quantitatively assess the information control capabilities of DisTIB as mentioned in \Cref{th: 1}.
In detail, we conduct experiments on MNIST as follows: 1) classification tasks on disentangled features; 2) quantify the information in representations, \ie, $I(X;A)$ and $I(X;Z)$. 
We also report the results of previous explicit disentanglement methods \cite{pan2021disentangled,dang2024disentangled,hadad2018two,mathieu2016disentangling} for a fair comparison.
\Cref{fig: MI quant} shows that all methods perform well on classification tasks, \ie, good classification performance on $A$ while nearly random guess on $Z$. Intuitively, this observation seemingly suggests that all approaches have acquired effective disentangled representations: label-related information exclusively contains aspects relevant to prediction (\ie, well-classified), whereas sample-specific information solely captures label-irrelevant elements (\ie, random guess).

However, when quantifying the information within the representations, it is evident that previous competitors struggle to regulate the information in their disentangled representations, leading to the leak of redundant input information into $A$.
Notably, though DisenIB and DisGenIB have all theoretically proved their disentanglement efficacy, our DisTIB exhibits greater information control ability. 
We attribute the performance disparity to the complex optimization strategies introduced by explicit disentanglement of DisenIB (adversarial training) and DisGenIB (objective factorization), leading to sub-optimal solutions.
In contrast, DisTIB with implicit disentanglement requires no complex optimization strategies and yields superior results: it not only precisely encodes label-related information but also captures more sample-exclusive information due to the sufficiency term.

\subsubsection{Influence of Disentanglement}
We propose a novel objective in \Cref{eq: origin}, including the conventional prediction term to supervise the extraction of label-related information, and the sufficiency and compression (disentanglement) terms that serve as regularizers. We study the influence of these regularizers as follows.

\paragraph{Quantitative} We conduct experiments on adversarial robustness and FSL. \Cref{tab: ablation loss} shows performance drops severely without both terms due to insufficient disentanglement. 
The performance greatly improves with the disentanglement term due to the better disentangled features. 
However, the performance gap still exists since only prediction and disentanglement terms cannot ensure optimal disentanglement.
Moreover, DisTIB loses end-to-end sample generation ability without sufficiency term. 
The sufficiency term cannot contribute to the disentanglement, which even slightly undermines performance due to the additional optimization.
The best results are achieved with both terms, demonstrating the efficacy of DisTIB in disentangling representations.

\begin{figure}
  \centering
  \includegraphics[width=0.8\linewidth]{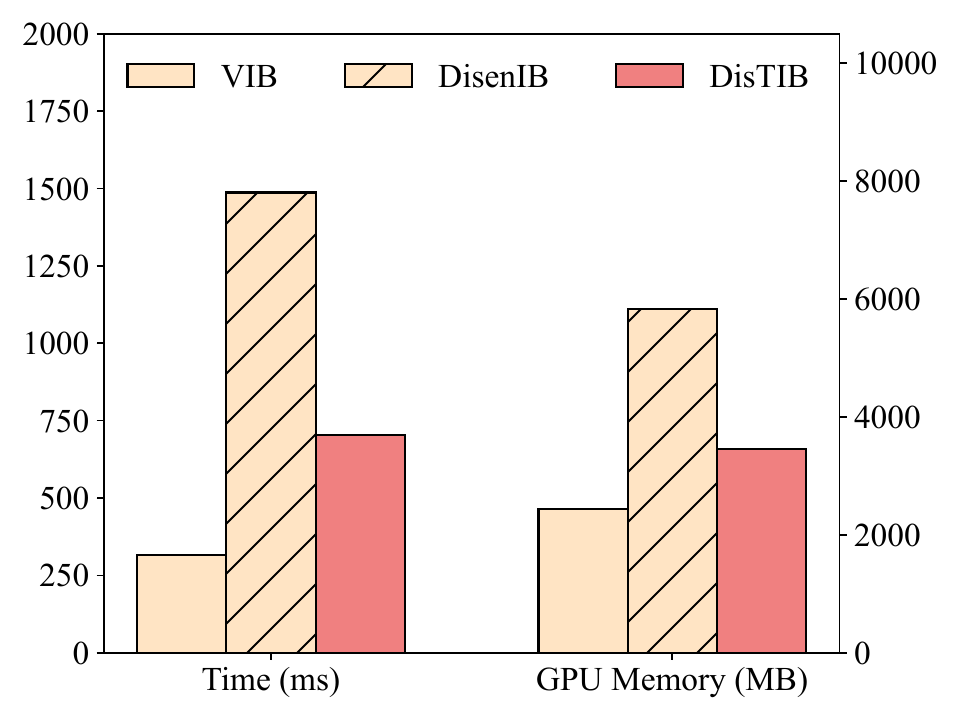}
  \caption{Per-batch average training time and GPU memory on MNIST.}
  \label{fig: time}
\end{figure}

\paragraph{Qualitative} We visualize prototypes and sample generation with/without disentanglement.
\Cref{fig: sprites_bad,fig: gen_bad} show that insufficient disentanglement seriously harms sample generation.
In contrast, in \Cref{fig: gen_good}, DisTIB can synthesize discriminative and diverse samples that lie coherently with the true labeled samples, effectively alleviating the data scarcity of FSL. 
Moreover, \Cref{fig: proto_vis} shows that prototypes generated by SG are more accurate than the conventional mean prototype, leading to more precise decision boundaries.

\subsubsection{Model Complexity}
Given input data dimension $N_D$, disentangled representation dimensions $N_A$ and $N_Z$ (assume $N_A = N_Z$), and the label dimension $N_Y$. Note that for a conventional information bottleneck baseline VIB \cite{alemi2016deep}, its optimization involves $I(X;A)-I(A;Y)$ with variational estimations, resulting in a training time complexity of $O(N_A \times (N_D + N_Y))$.
	Moreover, our DisTIB has a training time complexity of $O(N_A \times (4N_D + N_Y))$, which only involves an increase in constant terms and thus maintains comparable training time complexity with VIB.
	Furthermore, \Cref{fig: time} reports per batch training time and GPU memory on MNIST of our DisTIB and previous competitors to verify our analyses.
	Specifically, baseline VIB \cite{alemi2016deep} employs the simplest architecture and maintains the least time and memory cost.
	Differently, DisenIB \cite{pan2021disentangled} employs complex adversarial-based density estimation for disentanglement, significantly increasing model complexity \cite{creswell2018generative}. 
	Conversely, our DisTIB introduces an additional encoder to capture the sample-specific information and a generator for disentangled generation. \Cref{eq: origin} shows the optimization objective is reduced to conventional multi-information constraints with two extra terms compared to baseline, leading to a slight increase in time and memory cost.

\subsubsection{Hyperparameter Analyses}
\begin{figure}[!t]
  \centering
    \subfloat[$I(X;A)$ quantization.]{
        \includegraphics[width=0.24\textwidth]{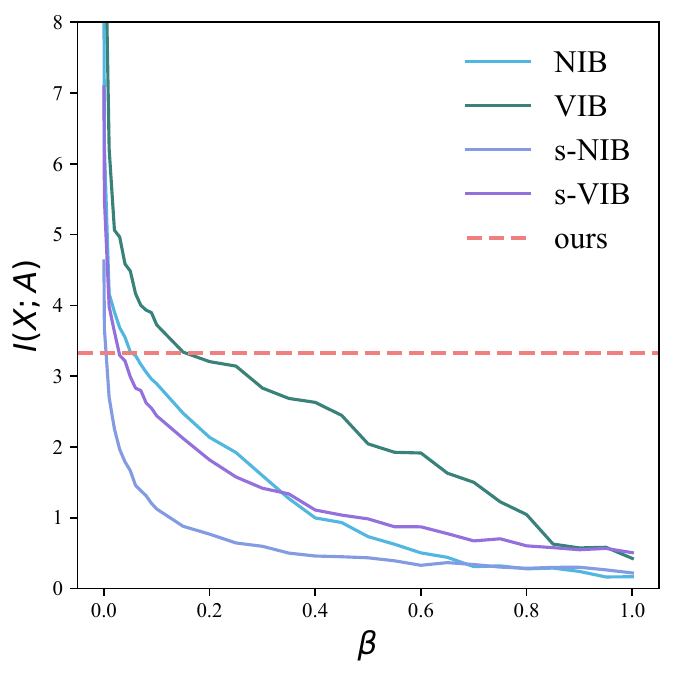}
    }
  \subfloat[$I(A;Y)$ quantization.]{
      \includegraphics[width=0.245\textwidth]{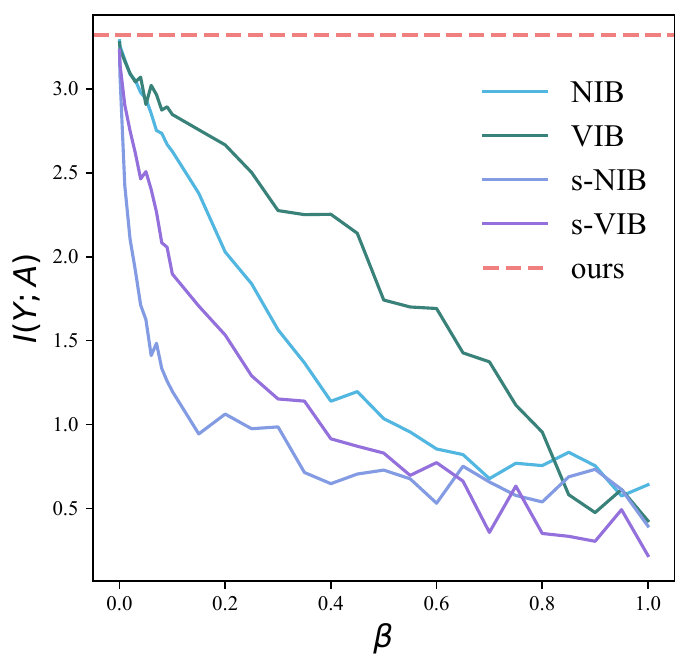}
  }
    \caption{
    The performance-$\beta$ and compression-$\beta$ trade-off on MNIST test set, where plots represent mutual information estimation; our DisTIB is represented as dotted lines due to its convergence to optimal disentanglement.}
    \label{fig.2}
\end{figure}

To study the influence of hyperparameter $\beta$, we first conduct experiments on MNIST with varying $\beta$.
In detail, we show the mutual information quantization with various values of $\beta$ in \Cref{fig.2}. We can observe a similar tendency on $\beta$ with previous works \cite{alemi2016deep,pan2021disentangled}, \ie, searching the performance-compression tradeoff inevitably degrades the performance of methods with trivial regularization. In contrast, our DisTIB exhibits a stable performance robust to $\beta$, attributed to its theoretically proven convergence to optimal disentanglement. These results are consistent with \Cref{fig: MI quant}, showcasing DisTIB's superior information control capability and its resulting enhanced performance-compression tradeoff.

\section{Conclusion and Discussion}
In this paper, we propose a novel transmitted information bottleneck for disentangling label-related and sample-exclusive information, named DisTIB. 
In detail, using variable interaction Bayesian networks, we extend the information bottleneck principle with transmitted information to formulate DisTIB. 
Moreover, we provide an in-depth theoretical analysis of DisTIB's disentanglement efficacy.
Experimental results reveal that DisTIB outperforms previous methods on various downstream tasks, confirming the validity of our theoretical analyses.
We hope our DisTIB offers new insights for future works to further enhance the disentanglement efficacy.

\noindent\textbf{Discussion and Future Work.}
We aim to highlight the critical role of data disentanglement in enhancing model generalizability, \eg, adversarial robustness, generalization and FSL. Our DisTIB demonstrates impressive empirical and theoretical disentanglement efficacy. More importantly, our DisTIB showcases full scalability as its core principle is input disentanglement independent of the dataset size.
Moving forward, we will explore new strategies and DisTIB's disentanglement efficacy on large datasets to enhance data efficiency.

\noindent\textbf{Limitations.}
Although DisTIB showcases impressive efficacy, it has several limitations. First, DisTIB implicitly assumes that every sample has correct labels for disentanglement, neglecting potential label noise in inputs and thus limiting its real-world applicability. Additionally, DisTIB relies on Bayesian networks to model variable interactions during training. However, in certain problems, the interactions between variables are complex and require expert knowledge to model accurately. We propose employing noise-robust strategies and causality mining methods to mitigate these issues.

    \bibliographystyle{IEEEtran}
    \bibliography{egbib}
    \vspace{-9mm}
    \begin{IEEEbiography}[{\includegraphics[width=1in,height=1.25in,clip,keepaspectratio]{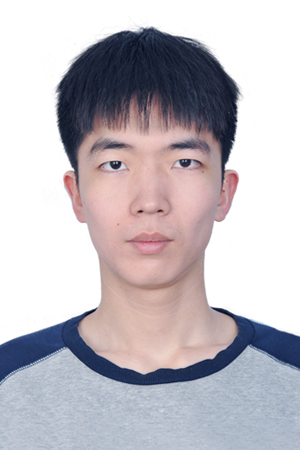}}]{Zhuohang Dang} received a BS degree from the Department of Computer Science and Technology, Xi'an Jiaotong University, in 2021. He is currently working toward the Ph. D. degree in Computer Science and Technology at Xi'''an Jiaotong University, supervised by Professor Minnan Luo. His research interests include causal inference and computer vision.
    \end{IEEEbiography}
    \vspace{-9mm}
      
      \begin{IEEEbiography}[{\includegraphics[width=1in,height=1.25in,clip,keepaspectratio]{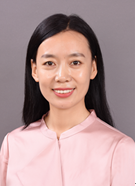}}]{Minnan Luo} received the Ph. D. degree from
      the Department of Computer Science and Technology, Tsinghua University, China, in 2014.
      Currently, she is a Professor in the
      School of Electronic and Information Engineering at Xi'an Jiaotong University. She was a PostDoctoral Research with the School of Computer
      Science, Carnegie Mellon University, Pittsburgh,
      PA, USA. Her research interests include machine learning and optimization, cross-media retrieval and fuzzy system.
      
      \end{IEEEbiography}
    \vspace{-9mm}

      \begin{IEEEbiography}[{\includegraphics[width=1in,height=1.25in,clip,keepaspectratio]{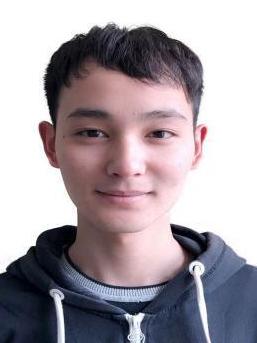}}]{Chengyou Jia} received the BS degree in Computer Science and Technology from Xi'an Jiaotong University in 2021. He is currently working toward the Ph. D. degree in Computer Science and Technology at Xi'an Jiaotong University. His
        research interests include machine learning and
        optimization, computer vision and multi-modal learning.
        \end{IEEEbiography}
    \vspace{-9mm}
      
        \begin{IEEEbiography}[{\includegraphics[width=1in,height=1.25in,clip,keepaspectratio]{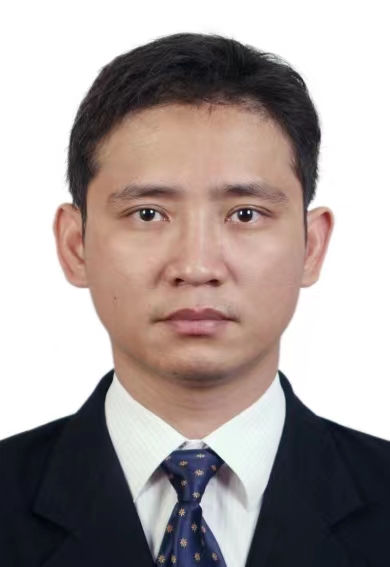}}]{Guang Dai} received his B.Eng. degree in Mechanical Engineering from Dalian University of Technology and M.Phil. degree in Computer Science from the Zhejiang University and the Hong Kong University of Science and Technology. He is currently a senior research scientist at State Grid Corporation of China. He has published a number of papers at prestigious journals and conferences, e.g., JMLR, AIJ, PR, NeurIPS, ICML, AISTATS, IJCAI, AAAI, ECML, CVPR, ECCV. His main research interests include Bayesian statistics, deep learning, reinforcement learning, and related applications.
        \end{IEEEbiography}
    \vspace{-9mm}
      
      \begin{IEEEbiography}[{\includegraphics[width=1in,height=1.25in,clip,keepaspectratio]{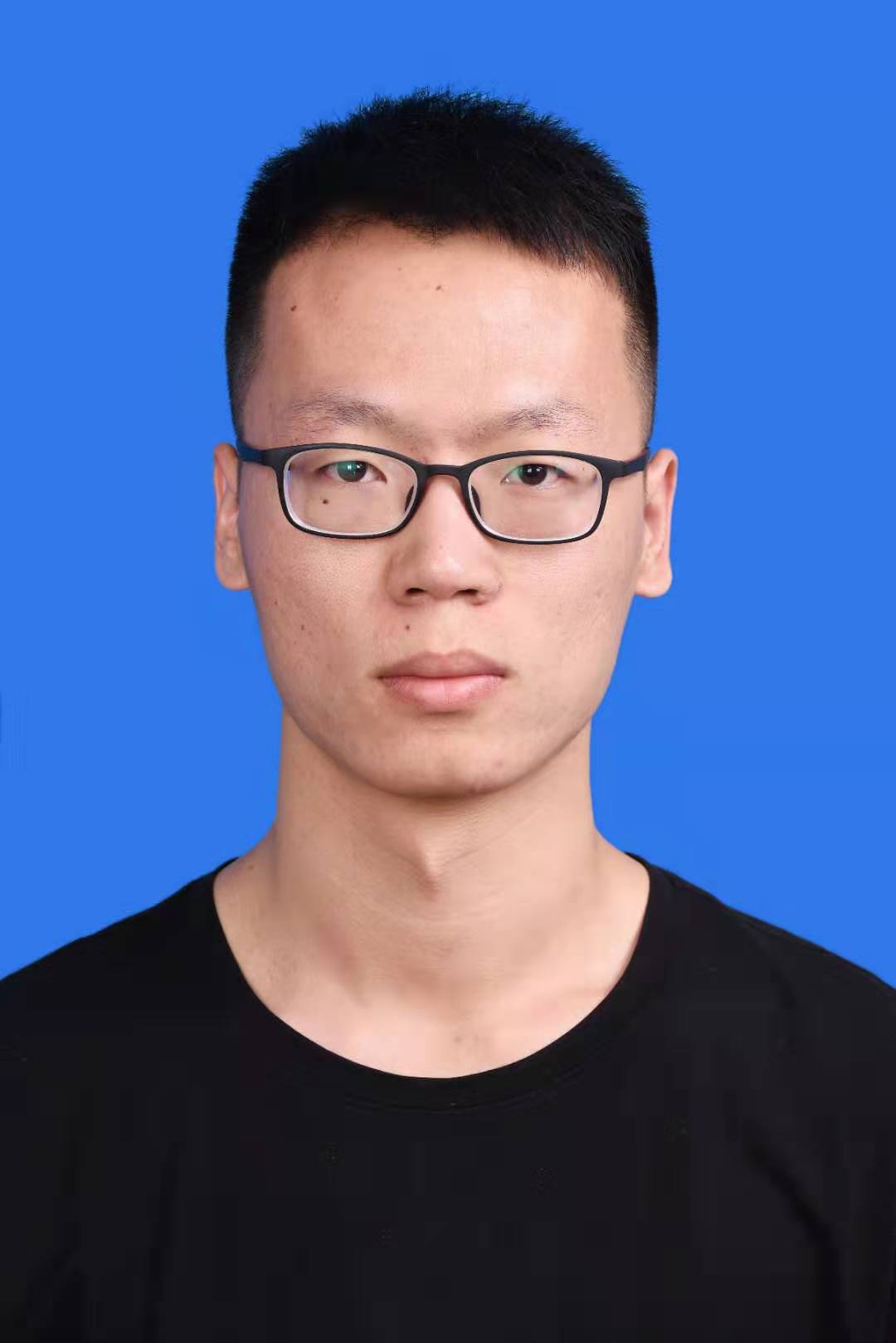}}]{Jihong Wang}
        received the B.Eng. degree from the School of Computer Science and Technology, Xi'an Jiaotong University, China, in 2019. He is currently a Ph.D. student in the School of Computer Science and Technology, Xi'an Jiaotong University, China. His research interests include robust machine learning and its applications, such as social computing and learning algorithms on graphs.
      \end{IEEEbiography}
    \vspace{-9mm}

      \begin{IEEEbiography}[{\includegraphics[width=1in,height=1.25in,clip,keepaspectratio]{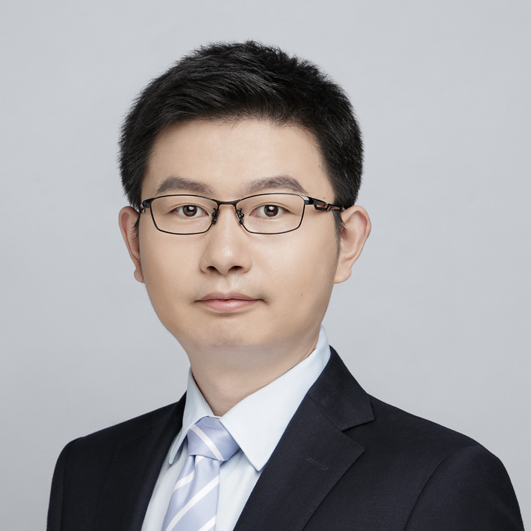}}]{Xiaojun Chang} is a Professor at Faculty of Engineering and Information Technology, University
      of Technology Sydney. He is also a visiting Professor at Department of Computer Vision, Mohamed bin Zayed University of Artificial Intelligence (MBZUAI). He was an ARC Discovery Early Career Researcher Award (DECRA)
      Fellow between 2019-2021. After graduation, he
      was worked as a Postdoc Research Associate
      in School of Computer Science, Carnegie Mellon University, a Senior Lecturer in Faculty of
      Information Technology, Monash University, and
      an Associate Professor in School of Computing Technologies, RMIT University. He mainly
      worked on exploring multiple signals for automatic content analysis in unconstrained or surveillance videos and has achieved top performance in various international competitions. He received his Ph.D. degree from University of Technology Sydney. His research focus in this period was mainly on developing machine learning algorithms and applying them to
      multimedia analysis and computer vision.
      \end{IEEEbiography}
    \vspace{-9mm}

      \begin{IEEEbiography}[{\includegraphics[width=1in,height=1.25in,clip,keepaspectratio]{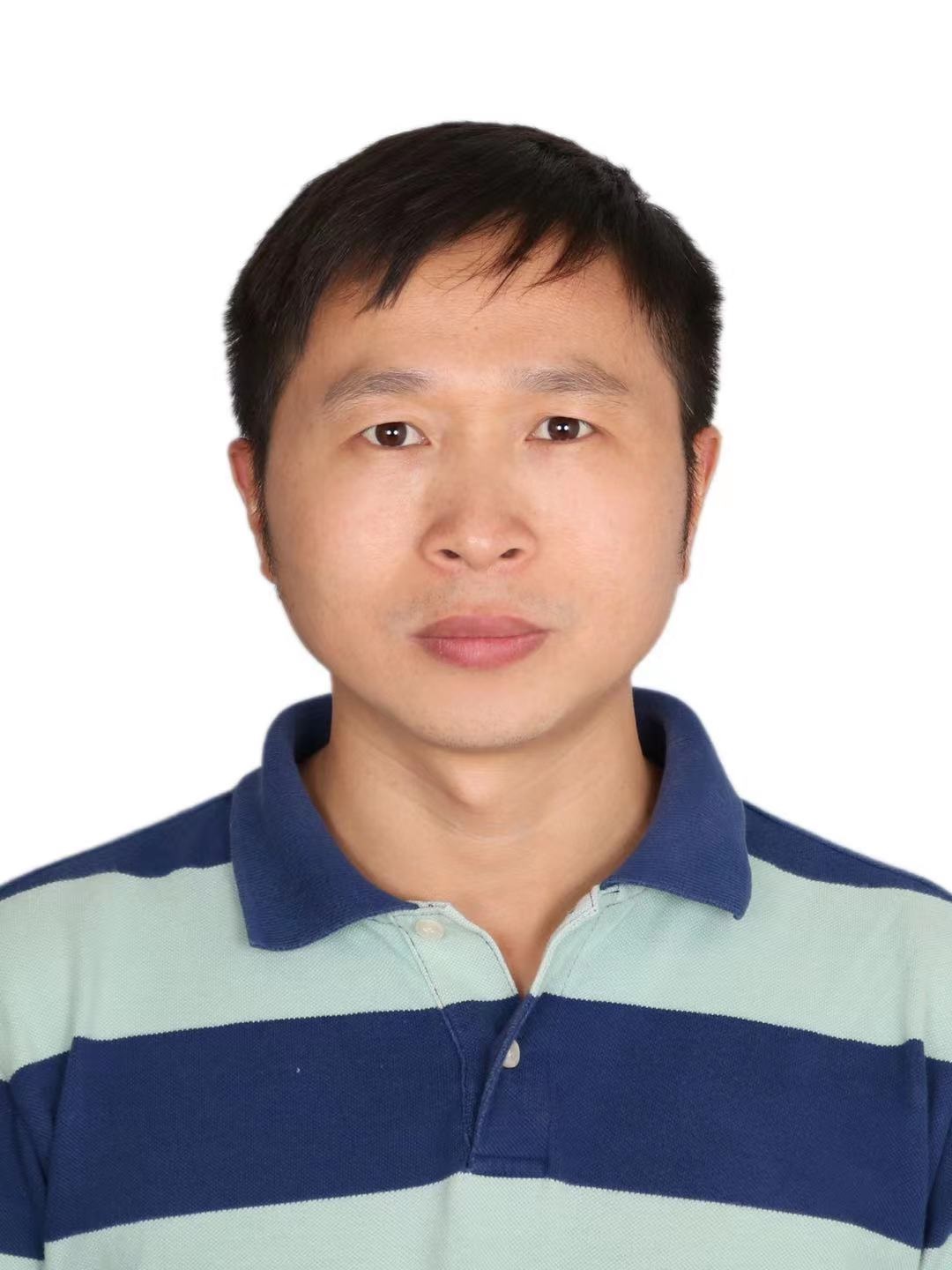}}]{Jingdong Wang} is Chief Scientist for computer vision with Baidu. Before joining Baidu, he was a Senior Principal Researcher at Microsoft Research Asia from September 2007 to August 2021. His areas of interest include vision foundation models, self-supervised pretraining, OCR, human pose estimation, semantic segmentation, image classification, object detection, and large-scale indexing. His representative works include high-resolution network (HRNet) for generic visual recognition, object-contextual representations (OCRNet) for semantic segmentation discriminative regional feature integration (DRFI) for saliency detection, neighborhood graph search (NGS, SPTAG) for vector search. He has been serving/served as an Associate Editor of IEEE TPAMI, IJCV, IEEE TMM, and IEEE TCSVT, and an (senior) area chair of leading conferences in vision, multimedia, and AI, such as CVPR, ICCV, ECCV, ACM MM, IJCAI, and AAAI. He was elected as an ACM Distinguished Member, a Fellow of IAPR, and a Fellow of IEEE, for his contributions to visual content understanding and retrieval.
      \end{IEEEbiography}
\end{document}